%% file: iccp23_template.tex
\documentclass[10pt,journal,compsoc]{IEEEtran}
\newif\ifpeerreview

\peerreviewfalse

\usepackage[nocompress]{cite}
\usepackage{url}
\usepackage{amsmath,amssymb,graphicx}

\usepackage{lipsum} %
\usepackage{enumitem}

\usepackage[switch]{lineno}

\newcommand{\paperID}{11}

\title{Measured Albedo in the Wild: Filling the Gap in Intrinsics Evaluation}

\author{Jiaye~Wu, 
Sanjoy~Chowdhury, 
Hariharmano~Shanmugaraja \\
David~Jacobs, 
and~Soumyadip~Sengupta
        
\IEEEcompsocitemizethanks{\IEEEcompsocthanksitem Jiaye Wu, Sanjoy Chowdhury, Hariharmano Shanmugaraja, David Jacobs are with the Department
of Computer Science, University of Maryland, College Park,
MD.
\IEEEcompsocthanksitem Soumyadip Sengupta is with the Department
of Computer Science, University of North Carolina, Chapel Hill,
NC.}%
}

\begin{document}

\IEEEtitleabstractindextext{%
\begin{abstract}
   Intrinsic image decomposition and inverse rendering are long-standing problems in computer vision. To evaluate albedo recovery, most algorithms report their quantitative performance with a mean Weighted Human Disagreement Rate (WHDR) metric on the IIW dataset. However, WHDR focuses only on relative albedo values and often fails to capture overall quality of the albedo. In order to comprehensively evaluate albedo, we collect a new dataset, Measured Albedo in the Wild (MAW), and propose three new metrics that complement WHDR: intensity, chromaticity and texture metrics. We show that existing algorithms often improve WHDR metric but perform poorly on other metrics. We then finetune different algorithms on our MAW dataset to significantly improve the quality of the reconstructed albedo both quantitatively and qualitatively. Since the proposed intensity, chromaticity, and texture metrics and the WHDR are all complementary we further introduce a relative performance measure that captures average performance. By analysing existing algorithms we show that there is significant room for improvement. Our dataset and evaluation metrics will enable researchers to develop algorithms that improve albedo reconstruction. Code and Data available at: \url{https://measuredalbedo.github.io/}
\end{abstract}

}

\ifpeerreview
\linenumbers \linenumbersep 15pt\relax 
\author{Paper ID \paperID\IEEEcompsocitemizethanks{\IEEEcompsocthanksitem This paper is under review for ICCP 2023 and the PAMI special issue on computational photography. Do not distribute.}}
\markboth{Anonymous ICCP 2023 submission ID \paperID}%
{}
\fi
\maketitle

\IEEEraisesectionheading{
  \section{Introduction}\label{sec:introduction}
}
\input{content/introduction}

\input{content/background}

\input{content/dataset}

\input{content/experiments}
\input{content/conclusion}

\ifpeerreview \else
\section*{Acknowledgments}
Jiaye Wu and David Jacobs were supported in part by the National Science Foundation under
grant no. IIS-1910132 and IIS-2213335.
\fi

\bibliographystyle{IEEEtran}
\bibliography{iccp23_template}

\ifpeerreview \else

\fi

\end{document}

%% file: content/introduction.tex
Single-image intrinsic image decomposition and inverse rendering have a long history in computer vision. In their seminal work, Barrow et al.\cite{1978intrinsics} defines intrinsic as ``range, orientation, reflectance, and incidental illumination''. However, most recent work \cite{bi_2015, bi_2018, usi3d, Zhou_2015_ICCV,Fan_2018_CVPR,cgintrinsics, direct_intrinsic, shapenet_intrinsics,survey} focuses on decomposing an image \(I\) into albedo \(A\) and shading \(S\), following the model \(I = A \cdot S\), where \(\cdot\) is element-wise multiplication. The closely related problem, inverse rendering, attempts to jointly infer material properties including albedo, geometry, and lighting from images\cite{sirfs, nir_2019, complex_indoor, inverserendernet, sir_3d}. Recovering scene properties from single images of indoor scenes is highly challenging and under-constrained due to complicated material, lighting, and geometry. Recently, significant progress has been made using deep learning and  large-scale photo-realistic synthetic datasets\cite{cgintrinsics, openrooms, interiornet18}. \cite{iiw, Zhou_2015_ICCV, relationship_mid_level, human_judgement_2015, bi_2015, Nestmeyer_2017, bi_2018, bigtime, Fan_2018_CVPR, cgintrinsics, nir_2019, glosh, usi3d, complex_indoor, sir_3d, paradigms, niidnet}.

\begin{figure}[t]
    \centering
    \includegraphics[width=\linewidth]{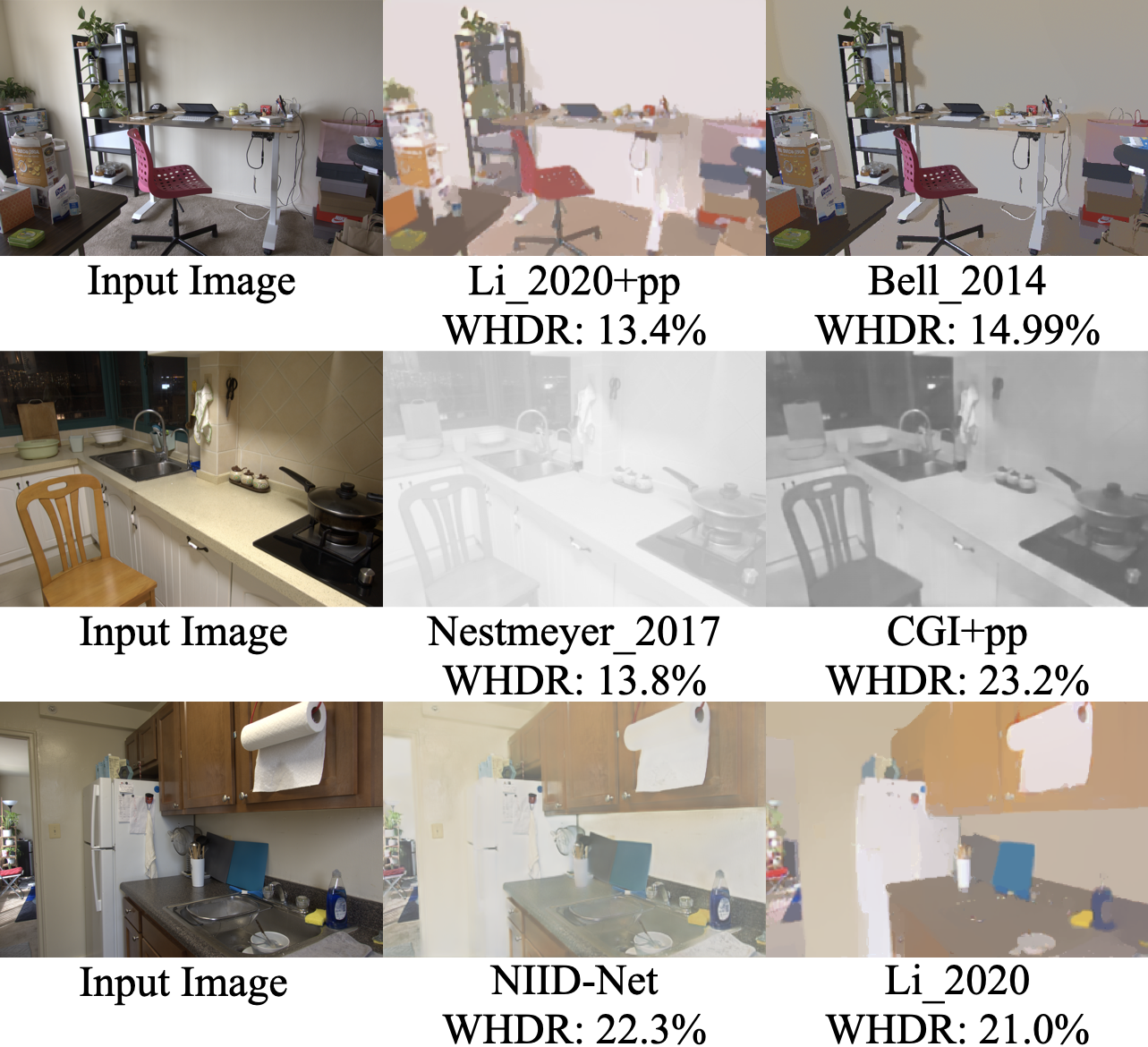}
    \caption{Better WHDR metric (lower is better) often do not produce better albedo, due to worse chromaticity (first row), intensity (second row), and texture (third row). Second row is shown in grayscale for best visualization of intensity. Additional discussions are in section \ref{sec:background}.}
    \label{fig:iiw_comparison_1}
\end{figure}
\begin{figure}[t]
    \centering
    \includegraphics[width=\linewidth]{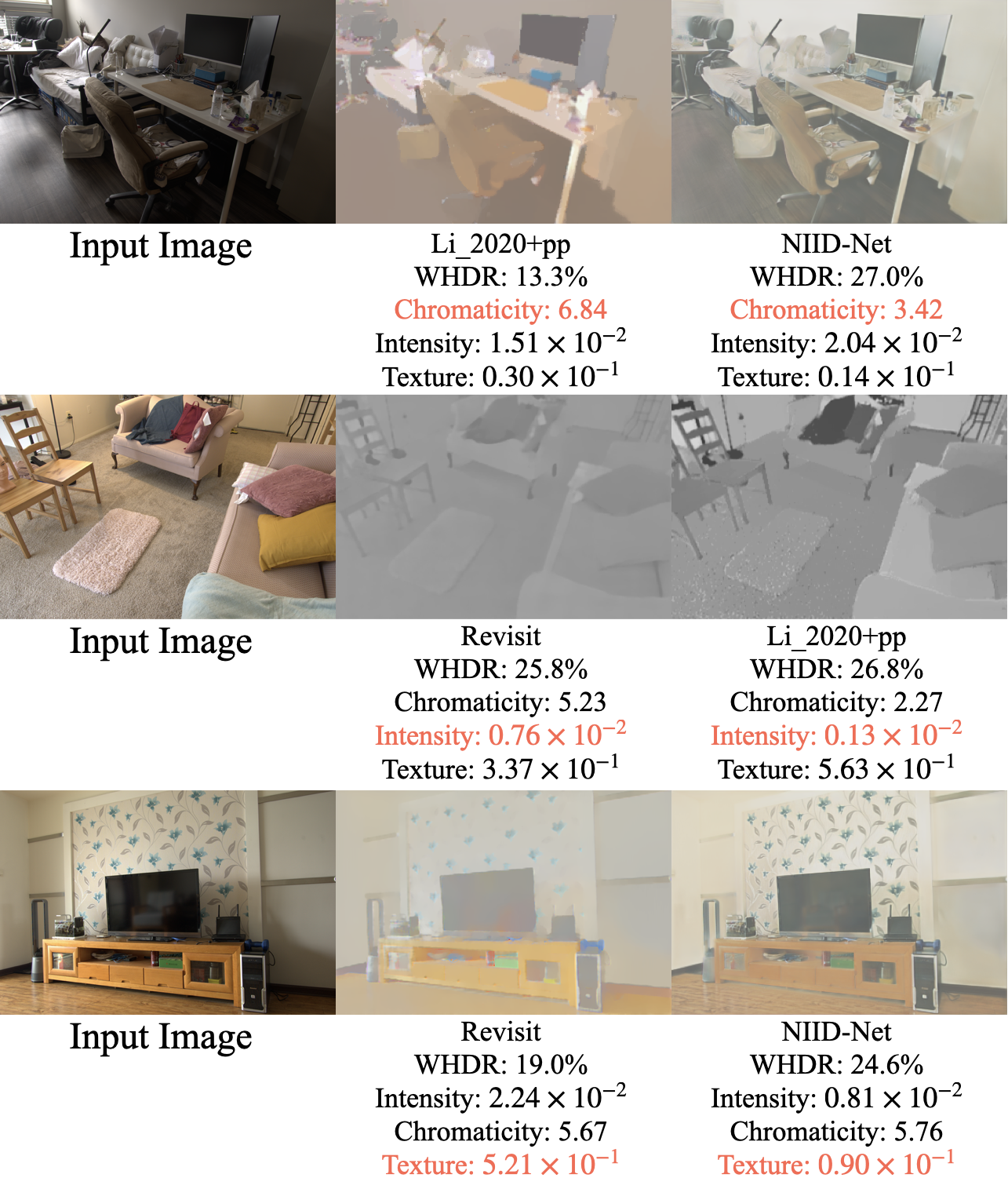}
    \caption{With all four metrics, we can correctly capture errors (lower is better) in chromaticity (first row), intensity (second row), and texture (third row). Second row is shown in grayscale for best visualization of intensity. Additional discussions are in section \ref{sec:testing}.}
    \label{fig:iiw_comparison_2}
\end{figure}

Progress in empirical research heavily relies on the availability of well constructed benchmarks to evaluate different algorithms. For the task of albedo prediction, for Intrinsic Image Decomposition and Inverse Rendering, the most widely used metric is the mean Weighted Human Disagreement Rate (WHDR) score on the IIW dataset~\cite{iiw} consisting of images of indoor scenes in-the-wild. WHDR score captures how well an algorithm agrees with human judgment of relative intensities in the albedo of a pair of points (darker, brighter, or equal). Despite its popularity, WHDR on IIW is incomplete. The authors of IIW \cite{iiw} themselves acknowledged that WHDR won't identify when colored shading is pushed into albedo or texture is interpreted as shading. Nevertheless, the authors found that ``algorithms that achieved lower WHDR also achieved qualitatively better results'' \cite{iiw}. Subsequent research \cite{paradigms} showed that some methods with good WHDR ``produce quite odd colors'', and ``display colored paper effect'', which is the artifact of a piece-wise constant albedo. While initially after the publication of IIW~\cite{iiw}, lower WHDR might predict qualitatively better results, we found researchers have since then overfit to the metric. As shown in Figure~\ref{fig:iiw_comparison_1}, algorithms with better WHDR metric often produce worse albedo reconstruction quality, especially in terms of recovering the correct intensity, chromaticity, and texture.

To address the limitations of the WHDR metric on the IIW dataset, we introduce a new benchmark for albedo estimation of indoor scenes in-the-wild. We present the Measured Albedo in the Wild (MAW) dataset, which contains 888 images of 46 indoor scenes with measured albedo and shading of 434 unique objects or regions. We introduce a novel and easy albedo measurement technique that allows a user to measure the albedo of an object in-the-wild scenes without requiring any expensive capture setup or constraints. Our method only requires a widely available gray card, a tripod, and a camera. This allows us to measure the albedo of most common objects, including curved objects, in a scene as long as they have a relatively small area (at least \(1'' \times 1''\)) that is flat, homogeneous, and approximately diffuse.  Ground truth shading images are derived from measured albedo. We then introduce several metrics to evaluate albedo including intensity, chromaticity and texture metrics which are complimentary to the WHDR metric. To evaluate the shading component, we use the same metric proposed in the MIT Intrinsics dataset \cite{mit_intrinsics}.

We first evaluate existing intrinsic decomposition or inverse rendering approaches on our proposed MAW dataset. Since these algorithms are only evaluated on WHDR metric for indoor scenes, the resulting algorithms often infer albedos that have low WHDR score but strong artifacts, as shown in Figure~\ref{fig:iiw_comparison_1}. However, the additional metrics introduced in this paper, related to intensity, chromaticity, and texture, can capture these artifacts in the inferred albedo as shown in Table~\ref{tab:metrics} and Figure~\ref{fig:iiw_comparison_2}.

  Our proposed MAW dataset and evaluation metrics can not only serve as a benchmark but also be used to train models to improve albedo predictions. We finetune various SOTA algorithms (Sengupta\_2019~\cite{nir_2019}, Li\_2020~\cite{complex_indoor}, and NIID-Net~\cite{niidnet}) on our MAW dataset using all four metrics (WHDR, intensity, chromaticity, and texture). We observe that finetuning significantly improves the predicted albedo and produces minimal artifacts, as shown in Table \ref{tab:finetuning} and Figure~\ref{fig:finetune_qual}. Specifically, finetuning improves the albedo in terms of intensity, chromaticity and texture metrics (36\%, 18\% and 8\% respectively on average) while slightly degrading WHDR (2\% on average). %

In summary, our contributions are as follows: 
\begin{itemize}[topsep=0pt]
    \item We introduce a new dataset called Measured Albedo in the Wild (MAW) that consists of 888 images of indoor scenes with ground truth albedo for evaluation. Shading can also be evaluated with shading annotation derived from measured albedo. 
    \item We show that although the WHDR metric provides a valuable way for evaluating albedo reconstruction, it is incomplete. Since it's introduction, researchers have overfitted to WHDR, improving WHDR at the cost of other aspects of albedo reconstruction, albedo intensity, chromaticity and texture. Our proposed MAW dataset including all four metrics allow us to properly evaluate the quality of the predicted albedo.
    \item We then show that finetuning various algorithms on our MAW dataset using all four albedo metrics significantly improves their overall performance both quantitatively and qualitatively. 
    \item Our results show that it is still an open problem to develop an algorithm that produces albedo that is good in all aspects, i.e. WHDR, intensity, chromaticity, and texture metrics. The dataset and evaluation code will be released upon acceptance. We believe this dataset can serve as a benchmark to future Intrinsic Image Decomposition and Inverse Rendering research.
\end{itemize}

%% file: content/background.tex
\section{Background} %
\label{sec:background}

Existing research made several attempts to evaluate albedo for intrinsic image decomposition and inverse rendering problems. Due to the lack of real datasets with high-quality albedo ground truth, many authors have evaluated their methods on synthetic data, such as MPI Sintel\cite{sintel}, and Interiornet\cite{interiornet18}. However, these do not fully capture the complex material, geometry and lighting of the real world. \textbf{MIT intrinsics}\cite{mit_intrinsics} is one of the first datasets with ground truth intrinsics for real objects (not scenes), measured in a controlled environment. \textbf{Intrinsic images in the wild} (IIW)~\cite{iiw} collects relative albedo annotations (darker, brighter, equal) for pairs of points for in-the-wild scenes through crowd-sourcing, and has become a de-facto standard for evaluating albedo reconstruction accuracy. However, such relative annotations cannot fully capture albedo reconstruction quality, see Figure~\ref{fig:iiw_comparison_1}. Our approach is a middle ground between MIT intrinsics and IIW. We measure groundtruth albedo like MIT intrinsics but for in-the-wild indoor scenes like IIW.

 \textbf{Intrinsic images in the wild.} Intrinsic images in the wild (IIW) \cite{iiw} is a large-scale dataset for evaluating albedo on indoor scene images captured in-the-wild. 106 pairs of points were randomly sampled on average for each image. Each pair of points is annotated by at least 5 MTurk workers who indicate which point has darker albedo (or that they are equal) with a confidence level. CUBAM \cite{cubam} is used to aggregate these annotations and confidence levels into a final relative judgement label (equal, point A darker, point B darker) and confidence score.

To quantitatively evaluate the performance of intrinsic image decomposition algorithms, IIW also proposes the weighted human disagreement rate (WHDR), which is the disagreement rate of the algorithm with humans, with the relative judgement weighted by its confidence score. Formally as defined in IIW \cite{iiw}: 
\begin{align} \small
    \operatorname{WHDR}_\delta(J, R) &= {\sum_i w_i \cdot \operatorname{\mathbf{1}}\left(J_i \neq \hat J_{i,\delta}(R) \right)}/{\sum_i w_i}, \nonumber
\end{align}
where $J_i$ is human relative judgement, $w_i$ is the confidence score, and $\operatorname{\mathbf{1}}(\cdot)$ is the unit indicator function. 

Since most intrinsic image decomposition algorithms output a 2D albedo image $R$ as the decomposed albedo, IIW also proposes a conversion function $\hat J_{i, \delta}$ that takes in $R$ and outputs the algorithm's judgement for the $i$'th pair. IIW \cite{iiw} defines the conversion function as follows:
\begin{equation} \small
    \hat J_{i, \delta} = \begin{cases}
        1 & \text{if } R_{2,i}/R_{1,i} > 1+\delta \\
        2 & \text{if } R_{1,i}/R_{2,i} > 1+\delta \\
        \text{E} & \text{otherwise}
    \end{cases}
\end{equation}
where $R_{1,i}$ and $R_{2,i}$ are the values at point A and point B in R respectively. $\delta$ is usually set to $10\%$.

 \textbf{WHDR metric on IIW.} Since its publication, WHDR on the IIW dataset has become the de-facto standard for quantitatively evaluating the albedo of intrinsic image decomposition and inverse rendering algorithms on real-world scenes. 
IIW\cite{iiw} itself includes an intrinsic image decomposition algorithm and evaluates the performance on their own dataset.
Bi\_2015\cite{bi_2015}, Zoran\_2015\cite{relationship_mid_level}, GloSH\cite{glosh}, Li\_2020\cite{complex_indoor}, Wang\_2021\cite{sir_3d}, Forsyth\_2021\cite{paradigms} and NIID-Net\cite{niidnet} does quantitative evaluation of albedo solely on WHDR on IIW.
Narihira\_2015\cite{human_judgement_2015} uses both WHDR and error rate (without weight) on IIW to evaluate their algorithm.
Bi\_2018\cite{bi_2018} conducts evaluation on IIW and the synthetic MPI-Sintel dataset \cite{sintel}. Also consistency of albedo under different illumination is evaluated on the \cite{bpb13} dataset.
BigTime \cite{bigtime} and CGIntrinsics\cite{cgintrinsics} evaluates on IIW, then quantitatively evaluates on MIT intrinsics dataset \cite{mit_intrinsics}.
Revisit\cite{Fan_2018_CVPR} and USI3D\cite{usi3d} evaluate on IIW for real indoor scenes, the MPI-Sintel dataset\cite{sintel}, and the MIT intrinsic dataset\cite{mit_intrinsics}.
Even though InverseRenderNet\cite{inverserendernet} is designed for outdoor scenes, the authors still decide to evaluate WHDR on IIW in addition to ``pseudo ground truth'' on megadepth\cite{megadepth}.
Direct Intrinsic \cite{direct_intrinsic} is designed for objects, but still reports WHDR on IIW.
IrisFormer\cite{irisformer} evaluates on IIW and a synthetic dataset interiornet\cite{interiornet18}.
Methods that didn't evaluate on IIW often use binocular stereo input \cite{lighthouse}, panoramic images \cite{ir_360}, or other forms of non-single-view input. Other methods that don't report WHDR scores include object-level methods \cite{shapenet_intrinsics}. %

 \textbf{Limitations of WHDR metric.} Limitations of the WHDR metric have been previously discussed in \cite{paradigms}. For example, in Figure \ref{fig:iiw_comparison_1} we show that often the quality of the reconstructed albedo is significantly better for the algorithm with the worse WHDR metric. In row 1, even though Li\_2020~\cite{complex_indoor} has lower WHDR, the wall is rendered pink and the chair red, which does not match the chromaticity of the input image. The chromaticity of Bell\_2014~\cite{iiw} is closer to the input image. On row 2, Nestmeyer\_2017~\cite{Nestmeyer_2017} has better WHDR score, but it is quite desaturated compared to CGI~\cite{cgintrinsics}, and does not have the correct albedo intensity value. On row 3, Li\_2020~\cite{complex_indoor} has slightly better WHDR score. However, the reconstructed albedo lacks the texture pattern on the kitchen counter and the wooden shelf. On the other hand, such texture patterns are much more well preserved in NIID-Net~\cite{niidnet}. In conclusion, we often notice that the WHDR measure does not correctly capture the intensity, color, and texture of the original albedo which is critical to evaluate albedo reconstruction quality.

%% file: content/dataset.tex
\begin{figure*}
    \centering
    \includegraphics[width=0.9\linewidth]{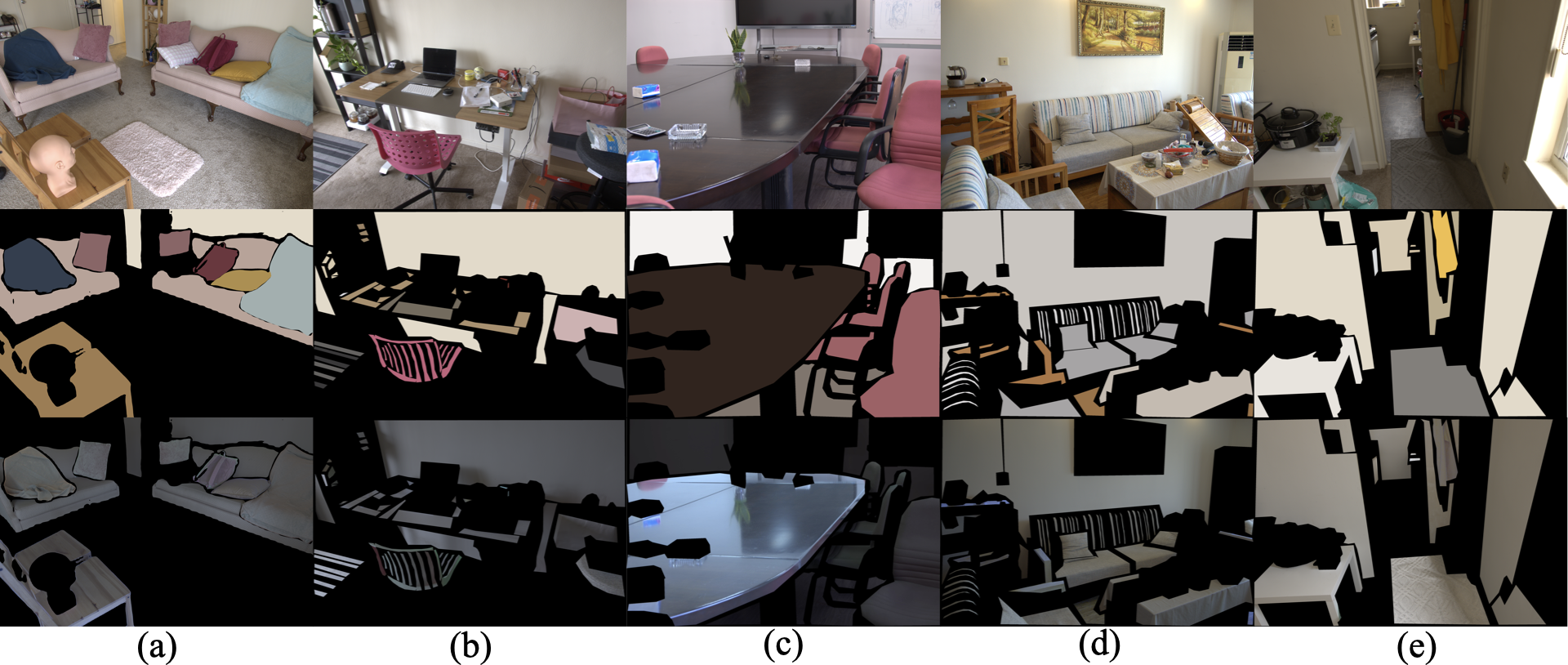}
    \caption{Images (top), ground-truth albedo (middle) and ground-truth shading from our MAW dataset.}
    \label{fig:albeo_measurements_qual}
\end{figure*}

\begin{figure*}[!t]
    \centering
    \includegraphics[width=0.9\linewidth]{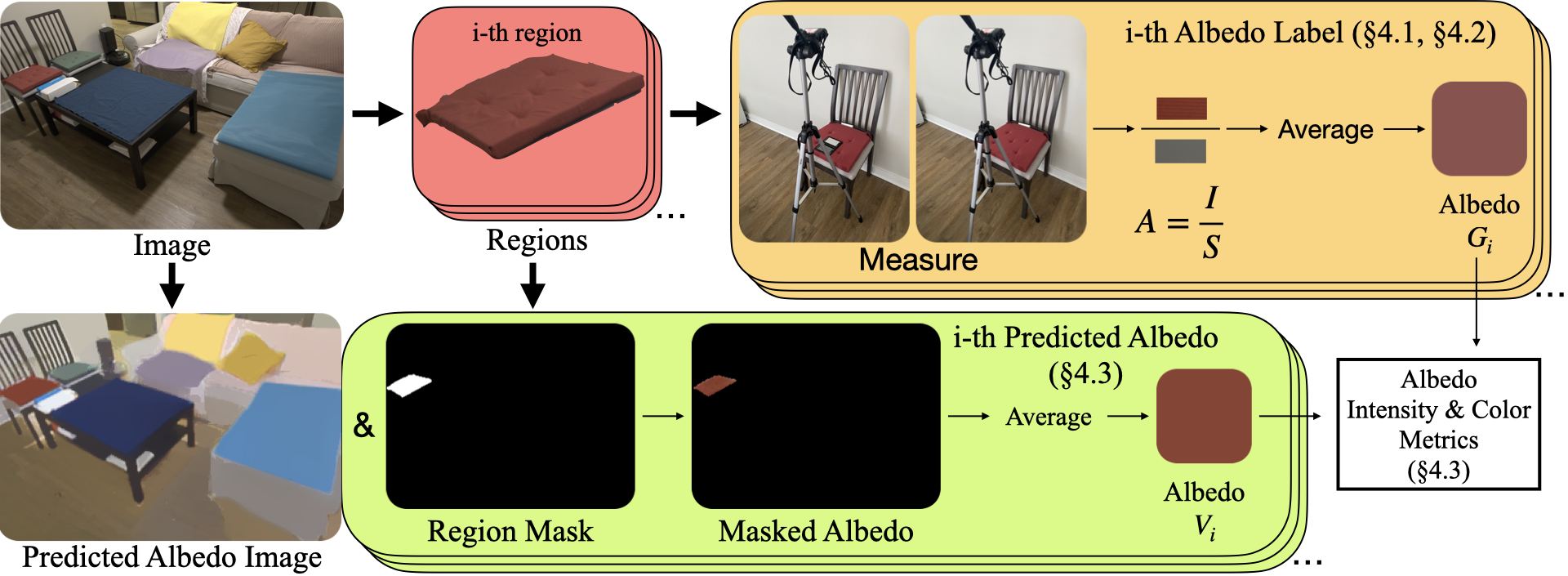}
    \caption{Overview of the data collection and evaluation pipeline of the MAW dataset. For each region with homogeneous texture, we measure a single average RGB value \(G_i\) for the entire region with a gray card. For a given predicted albedo image, we average pixels that lie inside the region as a single RGB value. We then compare the predicted albedo against the ground-truth using metrics as described in Sec. \ref{sec:albedo_metrics}. }
    \label{fig:measure_albedo}
\end{figure*}

\section{Measured Albedo in The Wild}

Accurately evaluating albedo intensity, chromaticity and texture require near ground-truth annotation of the albedo, which is nearly impossible to determine from just in-the-wild images. To the best of our knowledge, no such dataset exists for indoor scenes. Therefore, we collect such a dataset and introduce several metrics that capture the intensity, chromaticity and texture reconstruction quality. Our setup for capturing ground-truth albedo is lightweight and can be easily done in-the-wild, as we only need a camera, a gray card, and a tripod.

\subsection{Measuring Albedo}

\label{sec:measuring_albedo}
Capturing ground-truth albedo for all regions in a scene in the wild is extremely challenging. Thus we make the following assumptions, which hold true for many regions in images. We only measure albedo in these regions and evaluate different error metrics on these regions only.

\noindent$\bullet$ \textbf{Homogeneous.} We require the region to be \textit{homogeneous}. If the region contains texture, the feature size of the homogeneous texture to be significantly smaller than the size of a gray card (\(2.5'' \times 4''\)). We find that we can successfully measure most regions in a scene with such an assumption.

\noindent $\bullet$ \textbf{Not Highly Specular.} We avoid highly specular regions. In practice, we avoid specular reflections from a surface by capturing the surface from angles that minimize specularity. This allows us to capture reasonable albedo values for slightly specular objects (e.g., we can still capture a shiny table in Fig. ~\ref{fig:albeo_measurements_qual}(c)). We avoid highly specular surfaces due to concerns of specular reflections overpowering albedo.
  
\noindent $\bullet$ \textbf{Planar.} The region has an small \textit{planar} patch. We do not need the whole region to be planar as long as there is a small planar patch (significantly larger than the feature size of the homogeneous texture). Our method can measure the albedo of non-planar objects, e.g., the pink chair in Fig. \ref{fig:albeo_measurements_qual}(b) has curves but since the rim of the chair is flat we can provide measurements for the entire chair. Most common larger and salient curved objects in images such as bed, sofa, chair all have such local planar regions that we can measure.

To measure albedo of a region with unknown albedo and shading we use a `proxy object', which is a flat gray card, of known albedo value. We capture two images of the region, one by placing the `proxy object' on the surface of the region ($I_\text{proxy}$) and another without it ($I_\text{region}$). Since we ensure the planar geometry of both the region and the `proxy object', they will have the same shading value $S$. Since the albedo of the `proxy object' ($A_\text{proxy}$) is known, we can simply calculate the common shading $S=I_\text{proxy}/A_\text{proxy}$. Thus the albedo of the region can then be calculated as $A_\text{region} = I_\text{region}/S$.

Examples of image and measured albedo from the proposed MAW dataset are shown in Figure \ref{fig:albeo_measurements_qual}. We can measure a wide variety of objects, including sofa, chair, pillow, and even cylindrical plastic wash basins (by measuring the bottom). However, there are limitations to our method. We cannot measure non-homogeneous materials such as some carpet (Fig. \ref{fig:albeo_measurements_qual}(a)). We also cannot measure objects that don't have any planar region such as the flower pot (Fig. \ref{fig:albeo_measurements_qual}(b)). 

The previous albedo measurement benchmark, IIW \cite{iiw}, only provides relative comparisons of 44 pixels per image on average. In contrast we provide annotations on roughly 59,400 pixels per image at the same resolution as IIW.

\noindent\textbf{Details of Dataset Collection} The dataset collection is performed with a SONY RX-100 IV or SONY RX-100 VA or a NIKON D40 camera mounted on a tripod. For the gray card, we use either the 18\% gray card inside X-Rite/Calibrate ColorChecker Passport Photo 2 or a dedicated KODAK R-27 18\% gray card. When measuring albedo, all camera shots are taken with remote control in order to minimize misalignment between subsequent images.

To ensure image intensity of $I_\text{region}$ and $I_{proxy}$ increases linearly with respect to increasing amount of light reflected from $I_\text{region}$ and $I_\text{proxy}$, we use ``raw'' camera images to compute the $I_\text{region}$ and $I_{proxy}$, which has linear response after substracting the blackness level~\cite{hdrplus, cam_response}. We use opensource software rawpy~\cite{rawpy}, which is a python wrapper around the opensource raw processing library LibRaw~\cite{libraw}, to process the raw images into linear adobe RGB color space (adobe RGB color space without gamma curve). Since all algorithms implicitly assume the strictly smaller sRGB colorspace, we convert the computed linear adobe RGB color space albedo $A_\text{region}$ into sRGB color space for evaluation.

\subsection{Building Dataset of Measured Albedo}

\label{sec:data_collection}
An overview of how we collect our dataset and evaluate algorithms is shown in Figure \ref{fig:measure_albedo}. For each scene, we manually pick $N$ regions for which we will measure albedo. For each region, we measure the albedo of a patch using a gray card and use the average albedo of the patch as the albedo of the entire surface. Note that with our assumption, the texture feature size in the region being measured is significantly smaller than the gray card which varies in size from $2.5''\times 4''$ inches to $8'' \times 10''$. %

 To measure the albedo of the region, we move the camera on a tripod next to the surface and pick a direction that minimizes the specularity. With the camera held fixed, we take two images, with and without the gray card lying on the surface of the region being measured. We annotate the mask for the gray card. Since both images are in alignment, the mask allows us to crop out the gray card $I_\text{proxy}$ and surface under the gray card $I_\text{region}$ respectively, and we can compute the albedo $A_\text{region}$ as described in Section \ref{sec:measuring_albedo}. We take the average of \(A_\text{region}\) and denote as \(G_i\).

Then we capture $K$ images of a scene from different viewpoints, for some scenes under different lighting conditions too. Each of the $K$ images will contain only a subset of size $M$ of the $N$ measured surfaces, but from different viewpoint/lighting conditions. For each captured image, we manually create the segmentation mask of each visible surface as a collection of polygons or use the interactive segmentation tool F-BRS \cite{fbrs}. Then we manually associate each segmented region with one of the $M$ measured albedos \(G_i\).
In the end, we get albedo labels \(G_1, ... G_M\), and region masks \(M_1, ..., M_M\) associated with the image. 

In total we have $K=$ 888 images of 46 scenes, containing $N=$ 434 measurements. Of all possible image pairs, $93.5\%$ do not share any common measurements. Of those pairs that do share measurements, the IoU (Intersection over Union) is $0.316$.  Figure \ref{fig:albeo_measurements_qual} shows the variety of scenes in our dataset.

\subsection{Metrics for Albedo Reconstruction}

\label{sec:albedo_metrics}
In this section, we will introduce our metrics for albedo intensity and color. Unless otherwise specified, all RGB values used to compute our metrics are in linear RGB space without a gamma curve. Given a predicted albedo image $R$, with its j-th pixel denoted $R_j$ and binary masks $M_{i, j}$, which takes the value of 1 if pixel j belongs to region i, and 0 otherwise, we take the average value of predicted albedo inside each manually annotated region as the algorithm's prediction for that region. We compare $V_i = {\sum_j R_{j} M_{i, j}}/{\sum_j M_{i, j}}$ to ground truth with the following metrics.

\noindent \textbf{Albedo Intensity region-wise si-MSE} WHDR on IIW dataset uses the average of RGB channels as the albedo intensity values. Since most algorithms in intrinsic image decomposition follow this convention, we also use this method for computing the albedo intensity value. We denote grayscale predicted and ground truth albedo as $V^\text{gray}_i$ and $G^\text{gray}_i$ respectively.

We then compute the region-wise scale-invariant MSE between each prediction and the ground-truth region. Since the scales of albedos are ambiguous, we compute a single optimal global scale $\theta$ over the image that minimizes the sum of MSE between predicted and ground truth albedo values. We weigh each region by the number of pixels belonging to the region. Denote \(|M_i|\) as the number of valid pixels in region \(i\). Then region-wise si-MSE is computed with the global scale $\theta$ as:

\begin{equation} \small
   \hat{\theta} = \operatorname{argmin}_\theta \frac{ \sum_i |M_i| {\left\|V^\text{gray}_i - \theta G^\text{gray}_i\right\|}_2}{ \sum_i |M_i| },
\end{equation}

\begin{equation} \small
\label{eq:si-mse}
    \text{Intensity region-wise si-MSE} = \frac{ \sum_i |M_i| \|V^\text{gray}_i - \hat{\theta} G^\text{gray}_i\|_2^2}{\sum_i |M_i|}.
\end{equation}

\noindent \textbf{Albedo Chromaticity Error} For albedo chromaticity error, we want to completely ignore the grayscale component, and just capture the ``chromaticity'' component of the albedos. Therefore, instead of computing a single global scale $\theta$, we compute an optimal scale $\theta_i$ for each region i, such that with \(\theta_i\) applied, the grayscale predicted albedo \(\theta_i V_i^\text{gray}\) exactly matches groundtruth \(G_i^\text{gray}\). The difference between $V_i$ and $\theta_iG_i$ should now completely ignore the grayscale component and focus on the difference in chromaticity.

CIELAB \(\Delta E^*\) is a family of perceptually based metrics.  We use the latest CIEDE2000 version\cite{deltae2000} to measure the error in chromaticity. We weight each region by the number of pixels of that region. We denote CIEDE2000 as $\operatorname{\Delta E_{00}^*}(x, y)$, and \(|M_i|\) as the number of valid pixels in region \(i\).%

\begin{equation} \small
\label{eq:color}
    \text{Color Error} = \frac{\sum_i |M_i| \operatorname{\Delta E_{00}^*}\left(V_i, \theta_i G_i\right)}{\sum_i |M_i|}
\end{equation}

\label{sec:texture_detail}
\begin{figure}
    \centering
    \includegraphics[width=0.9\linewidth]{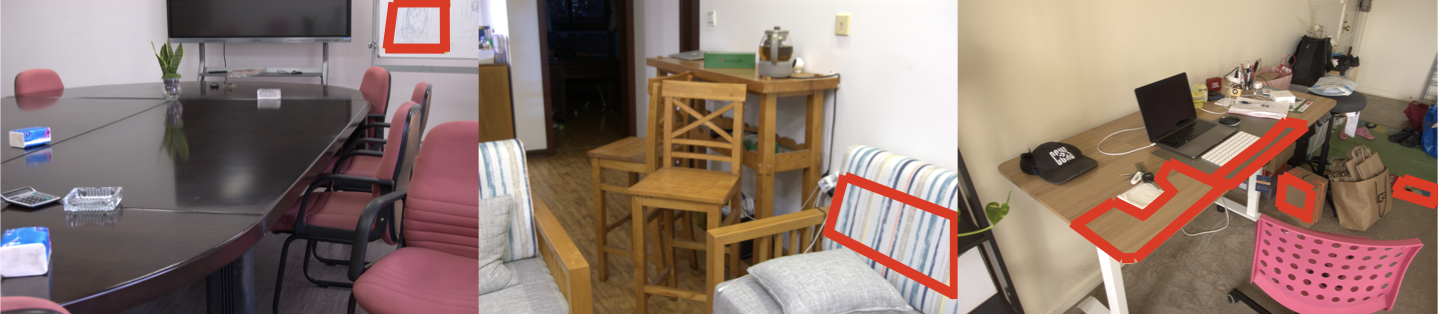}
    \caption{Approximately constant shading masks used for texture metrics. Red polygons indicates the locations of the constant shading masks. }
    \label{fig:texture_masks}
\end{figure}

\noindent \textbf{Albedo Texture Error} We propose a metric to measure how well texture details in the input image are preserved in the albedo. We know that in the regions of constant shading of an image, the ground truth albedo and the image only differ by a constant RGB factor. We thus rescale the predicted albedo in these regions such that the average RGB values match the average RGB values of the input image. Then we can use LPIPS\cite{lpips}, a metric of perceptual similarity, to evaluate how well the texture details of the input images are preserved in the predicted albedo in these regions.

We manually annotate the MAW dataset with regions of approximately constant shading. Some examples of constant shading masks are shown in Figure \ref{fig:texture_masks}. We extract the largest inscribed rectangles from the polygon annotations, and only keep rectangles with length and width at least 32. We upsample both the input image and predicted albedo to prevent too many rectangles from getting removed. 639 images from MAW has at least 1 rectangle matches the criteria and are used for texture evaluation.

\noindent \textbf{Albedo WHDR} Our albedo intensity and chromaticity metrics mainly focus on the absolute values of albedo prediction. We still use WHDR metric to capture relative reflectance between pairs of pixels. We collect WHDR label on MAW dataset with official code from IIW~\cite{iiw}. Annotating images with WHDR score is an economically expensive process. Thus we have only annotated WHDR labels for 20\% of the MAW dataset.

\subsection{Combining Different Performance Metrics}
Our proposed MAW dataset, contains ground-truth annotations to measure albedo intensity, chromaticity, WHDR scores, and texture reconstruction. These four metrics are complementary and capture different aspects of albedo reconstruction. However in practice, we observe that certain algorithms are better in some metrics and worse in others (e.g. Figure \ref{fig:iiw_comparison_2}, \ref{fig:mse_whdr_qual}, \ref{fig:texture_qual}). This leads to a key question: how do we know which algorithm performs better overall?

Proposing a combined metric is challenging as we do not know how to weight each metric. Even if we assume equal weights on all metrics, they are of very different scale and the metric with highest magnitude will dominate. Instead, we propose to measure the performance of any algorithm as an average relative improvement over all competing algorithms in all four metrics.

Given any two algorithms and their performance on a set of metrics, we can calculate the relative improvement of one algorithm over another in each of the metrics. Since relative improvement measures for each metric are in percentage we can combine them assuming equal weights. This indicates how much on average one algorithm improves over another by equally weighting all the metrics. In practice, we are often comparing one algorithm with many existing state-of-the-art algorithms. In this case, we simply take an average of the relative improvement of the given algorithm over all other algorithms.

Formally, let $A_i(m_j)$ denote the performance of algorithm $A_i$ on metric $m_j$ (lower is better). Then the overall relative improvement performance of algorithm $A_i$ over all other algorithms and metrics can be defined as:
\begin{align} \small
    R_{i,k}(m_j)&= (A_k(m_j)-A_i(m_j))*(\dfrac{1}{A_i(m_j)}+\dfrac{1}{A_k(m_j)}) \nonumber\\
    P(A_i) &= \dfrac{1}{L-1}\sum_{\substack{k=1 \\ k \neq i}}^{L} \dfrac{1}{M}\sum_{j=1}^{M} R_{i,k}(m_j),
    \label{eq:comb}
\end{align}
where M is the number of metrics and $L-1$ the total number of algorithms to be compared with. $R_{i,k}(m_j)$ is the relative improvement of algorithm $A_i$ over $A_k$ on metric $m_j$.

\subsection{Deriving Ground Truth Shading}
\label{sec:deriving_shading}

\begin{figure*}
    \centering
    \includegraphics[width=0.9\linewidth]{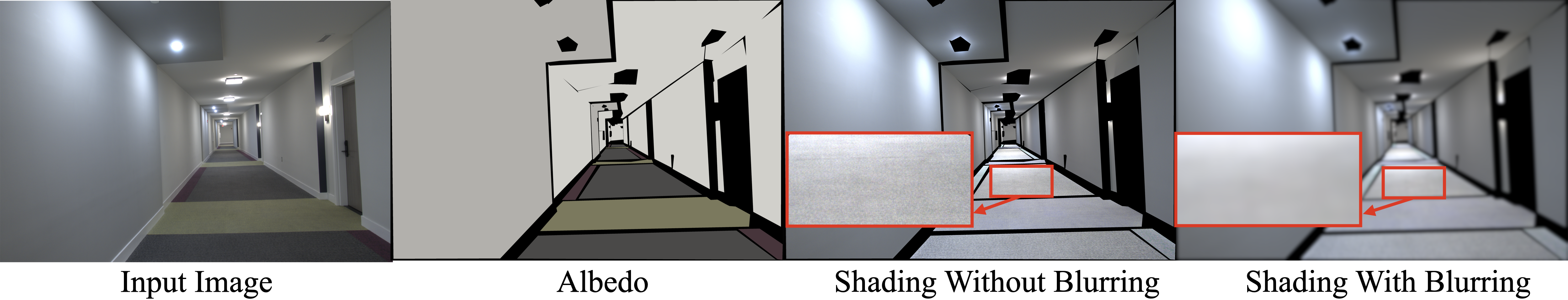}
    \caption{Shading annotation in the MAW dataset. Input image, albedo, derived shading without blurring, derived shading with blurring are shown respectively. Blurring the image allows deriving the smooth shading without the leaked texture details. As highlighted by the red box, the texture details on the carpet are leaked into shading without blurring, and blurring removed such leaked texture.}
    \label{fig:filter_shading}
    \vspace{-1.0em}
\end{figure*}
Given the intrinsic decomposition formulation of \(I=S*A\), we can easily derive shading as \(S=I/A\). However, many regions are annotated with homogeneous albedo without texture details, such as the carpet in Figure \ref{fig:filter_shading}. In these cases, high frequency texture details will leak into derived shading \(S\) after the division. As our albedo is approximately homogeneous, we treat the leaked texture detail as high frequency noise, and blur the derived shading image, which will produce a blurred shading image \(\hat{S}\) without the leaked texture details. We blur the shading image with a gaussian filter wth standard deviation of \(\sigma\). i.e. \(\hat{S}=\operatorname{F}_\sigma(S)\). As shown in Figure \ref{fig:filter_shading}, without the gaussian blurring, texture leaks into the carpet, which is removed with gaussian blurring.

\subsection{Metric for Shading}
\begin{figure*}
    \centering
    \includegraphics[width=0.9\linewidth]{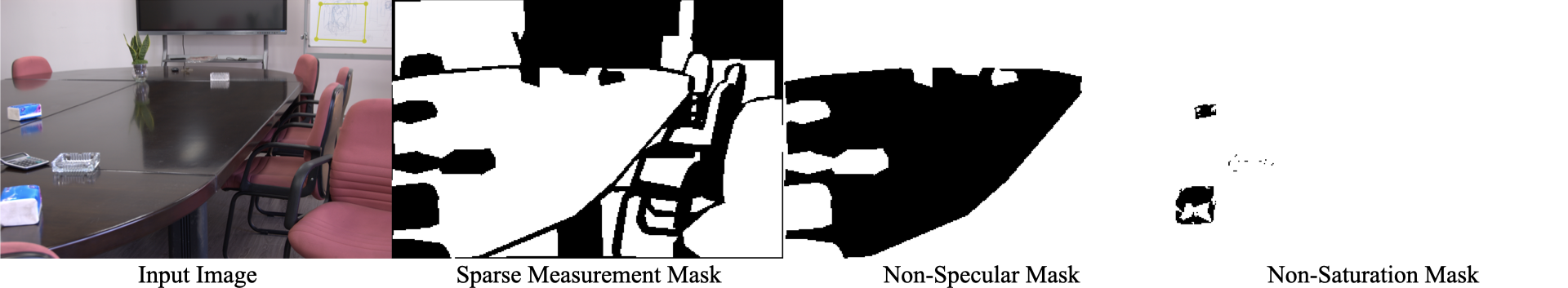}
    \caption{Sparse measurement mask, non-specular mask, and non-saturation mask for shading metrics.}
    \label{fig:shading_masks}
\end{figure*}
We can evaluate algorithms with our derived groundtruth shading. Although we can just apply si-MSE loss from MIT Intrinsics\cite{mit_intrinsics},  our derived groundtruth shading is blurred to remove leaked texture. To ensure a fair benchmark, we apply the blurring operation to the predicted shading to obtain blurred predicted shading \(\hat{S}_\text{pred} = F_\sigma (S_\text{pred})\).%

Our measurements might be sparse, contain specularities, and can overexpose, and we want to ignore some pixels. We introduce sparse si-MSE. Given blurred groundtruth and predicted shading \(\hat{S}_\text{gt}\) and \(\hat{S}_\text{pred}\), and the mask for shading \(M_\text{shading}\), the sparse si-MSE is defined as the following, where \(\hat{\theta}\) is the optimal \(\theta\) that minimizes the equation:
\begin{equation}
    \text{sparse si-MSE} = \frac{\|(\hat{S}_\text{gt} -  \hat{\theta}\hat{S}_\text{pred}) \odot M_\text{shading}\|}{\sum_i\sum_j M_{\text{shading}, i,j}}
\end{equation}
Our shading mask \(M_\text{shading}\) is the intersection of the Sparse Measurement Mask (\(M_\text{sparse}\)), Non-Specular Mask (\(M_\text{non-specular}\)),  and Non-Saturation Mask (\(M_\text{non-saturation}\)), which are introduced below. Figure \ref{fig:shading_masks} show examples of such masks.

\noindent\textbf{Sparse Measurement Mask} (\(M_\text{sparse}\)) We set the mask to 1 only if there is valid shading at the pixel.

\noindent\textbf{Non-Specular Mask} (\(M_\text{non-specular}\)) There does not seem to be a single convention for handling specular regions in intrinsic decomposition. While many works \cite{iiw, Nestmeyer_2017, cgintrinsics, Fan_2018_CVPR, usi3d, niidnet} leave specularity in the shading image with \(I=A*S\), thus \(S=I/A\), other works \cite{shapenet_intrinsics, Baslamisli_2018} assume that there is a separate specularity channel \(I=A*S+R\), where \(R\) is the additive specular component. Moreover, a practical issue arises from specular regions when we try to evaluate shading decomposition. A glossy surface with very dark albedo will result in huge derived shading value \(S=I/A\), since \(I\) is large due to specular reflection and \(A\) is small as the surface is dark. One such example is the glossy table in Figure \ref{fig:albeo_measurements_qual} (c). Therefore, we manually annotate glossy regions in our dataset. We set the non-specular mask to 1 only if the pixel is not inside any glossy region.

\noindent\textbf{Non-Saturation Mask} (\(M_\text{non-saturation}\)) To avoid using shading derived from over-saturated pixels, we set non-saturation mask to 1 only if the pixel value is less than 250 out of 255 in the input image \(I\).

\noindent\textbf{Combined Shading Metric Mask} (\(M_\text{shading}\)) Final combined mask is: \(M_\text{shading}=\bigcap (M_\text{sparse}, M_\text{non-specular}, M_\text{non-saturation})\).

\section{Finetuning with MAW dataset}

\label{sec:train}

While MAW can serve as a benchmark for evaluating albedo and shading predictions, we show that it is also possible to use MAW to finetune various intrinsic image and inverse rendering algorithms and thus improve its performance. Although MAW has only 888 images, it is still very effective to improve the performance of various algorithms which are trained on larger datasets (synthetic and/or IIW).

Denote \(G_i\) as the scalar RGB value for the i-th region, \(R\) as the predicted albedo. \(M\) is a mask where \(M_{i, j}\) takes value of 0 if pixel j belongs to region i and 0 otherwise. We compute the global scale $\theta$ that minimizes the loss. Then the si-MSE is computed.
\begin{equation} \small
     \operatorname{si-MSE Loss}(R, G, M) = \frac{\sum_i \sum_j \|(\hat{\theta} M_{i,j} R_j - G_i)\|_2^2}{\sum_i \sum_j M_{i,j}}.
\end{equation}

Since si-MSE does not capture the relative relationship between pixels, training on si-MSE loss might degrade performance on the WHDR metric. Therefore, we also finetune with WHDR labels on the MAW dataset with the hinge proxy loss for WHDR metric defined in CGI\cite{cgintrinsics} as:
\begin{align}
    &\operatorname{H}_\tau(R_i, R_j, W_{i,j}, J_{i,j}) \\
    &=\begin{cases}
        \alpha_{+} w_{i,j}(R_i - R_j)^2 & \text{if } J_{i,j} = 0 \\
        \alpha_{-} w_{i,j}(\operatorname{max}(0, \tau - R_i + R_j))^2 & \text{if } J_{i,j} = +1 \\
        \alpha_{-} w_{i,j}(\operatorname{max}(0, \tau - R_i + R_j))^2 & \text{if } J_{i,j} = -1 \\
    \end{cases}
\end{align}
where $R_i$ and $R_j$, are pixel $i$ and $j$ of the predicted albedo. $w_{i,j}$ is the confidence score that comes with WHDR annotations, and $J_{i, j}$ is the  judgement label between pixel $i$ and $j$. $-1$ indicates point $i$ is darker, and $+1$ indicates point $j$ is darker. $\alpha_{+}$ and $\alpha_{-}$ are our weight for equal and not equal judgements respectively. We set $\alpha_{+}$ to $\frac{1}{|\text{number of equal judgements}|}$ and we set $\alpha_{-}$ to $\frac{1}{|\text{number of unequal judgements}|}$. The final hinge loss is $\operatorname{H}_\tau(R, W, J)=\sum_{i,j} \mathbf{1}(J_{i,j}) \operatorname{H}_\tau(R_i, R_j, W_{i,j}, J_{i,j})$, where \(\mathbf{1}(J_{i,j})\) is 1 if \(J_{i, j}\) exists, 0 otherwise. We set \(\tau\) to 0.2 in our experiments.

Texture error for training is similar to our texture metric for the benchmark, which we denote as \(\operatorname{Tex}(R, I, M_\text{shading})\), where \(I\) is the image, and \(M_\text{shading}\) the shading metric mask.

We use our collected WHDR judgements to finetune various algorithms with IIW\cite{iiw}. Even though WHDR judgement is only annotated for 20\% of the images, we found that this is sufficient to fine-tune and improve the overall performance of different algorithms. Therefore, our final combined loss for each image is:
\begin{align} 
    \operatorname{L_{\text{final}}}&(I, R, G, M, J, M_\text{shading})=\\
    &\operatorname{si-MSE Loss}(R, G, M)
    + \beta \operatorname{\mathbf{1}}(H_\tau(R, W, J))\\
    &+ \gamma \operatorname{Tex}(R, I, M_\text{shading})
\end{align}
where $\operatorname{\mathbf{1}}(H)$ is equal to hinge loss when the image contains WHDR annotation and 0 otherwise. In our experiments, we set \(\beta\) to 2. We set \(\gamma\) to \(0\) for Li\_2020\cite{complex_indoor} and Sengupta\_2019\cite{nir_2019} because their texture does not degrade without \(\operatorname{Tex}\). We set \(\gamma\) to \(0.0005\) for NIID-Net\cite{niidnet}. We show in Table \ref{tab:finetuning} that such finetuning significantly improves the overall performance, in particular, albedo intensity and chromaticity, while WHDR might degrade marginally.

%% file: content/experiments.tex
\section{Results and Discussions} %

\begin{table*}[t]
    \centering
    \caption{ We report albedo intensity, (\(M_\text{saturation}\))chromaticity, WHDR, texture and shading score on our MAW dataset, and average Relative Improvement over all algorithms on all albedo metrics (which excludes shading metric). ``+pp'' indicates post-processing proposed by respective original algorithms. Rows are sorted according to WHDR metric. (Only ``+pp'' variant is included in sorting if both are present.) Some algorithms do not predict color but rather use the input image chromaticity, their chromaticity is indicated by *. Sengupta\_2019 and Li\_2020 are inverse rendering algorithms and do not predict shading directly. Their shading is derived from image divided by their predicted albedo, which is indicated by *.}
    \scalebox{0.91}{

    \begin{tabular}{|c|c|c|c|c|c|c|}
    \hline
    {} & {WHDR (\%)} & {Intensity (\(\times 10^{-2}\))} & {Chromaticity}  & {Texture(\(\times 10^{-1}\))} & {Rel. Improv. (\%)} & {Shading Score}\\
    {} & Lower is better ($\downarrow$) & Lower is better ($\downarrow$) & Lower is better ($\downarrow$) & Lower is better ($\downarrow$) & Higher is better ($\uparrow$) & Lower is better ($\downarrow$)\\
   \hline
   {Revisit \cite{Fan_2018_CVPR}} & 20.0 & 3.40 & 6.56* & 2.38  & - & 0.107 \\
   \hline
    {Revisit \cite{Fan_2018_CVPR}+pp} & 19.5 & 3.46 & 6.56* & 2.63 & -36.1 & - \\
    \hline
   {Li\_2020 \cite{complex_indoor}} & 31.0 & 1.36 & 5.70 & 2.43 & - & 0.094*\\
   \hline
    {Li\_2020 \cite{complex_indoor}+pp} & 19.8 & 1.41  & 5.64  & 2.52 & +29.3 & - \\
    \hline
   {CGI \cite{cgintrinsics}} & 26.1 & 1.72 & 6.56*   & 1.93 & - & 0.108\\
   \hline
   {CGI \cite{cgintrinsics}+pp} & 20.4 & 1.81 & 6.56*  & 1.98 & +17.0 & - \\
   \hline
   {Sengupta\_2019 \cite{nir_2019}} & 22.3 & 2.17 & 6.39 & 2.42 & -17.2 & 0.120* \\
      \hline
    {Nestmeyer\_2017\cite{Nestmeyer_2017}} & 23.7 & 3.60 & 6.56* & 1.72 & - & 0.112\\
   \hline
   {Nestmeyer\_2017\cite{Nestmeyer_2017}+pp} & 22.9 & 3.62  & 6.56* & 2.03  & -33.7 & - \\
   \hline
   {Bell\_2014\cite{iiw}} & 23.1 & 3.11 & 6.61 & 1.49 & -6.7 & 0.112\\
   \hline
   {NIID-Net \cite{niidnet}} & 25.5 & 1.24 & 4.73 & 1.22 & +77.1 & 0.082 \\
   \hline
    {BigTime \cite{bigtime}} & 28.5 & 2.71 & 5.15 & 2.14 & -9.3 & 0.133 \\
   \hline
    {USI3D \cite{usi3d}} & 29.5 & 2.62 & 6.00  & 1.93 & -19.9 & 0.115 \\
    \hline

    \end{tabular}
} %

    \label{tab:metrics}
  \end{table*}

\begin{figure*}
    \centering
    \includegraphics[width=\textwidth]{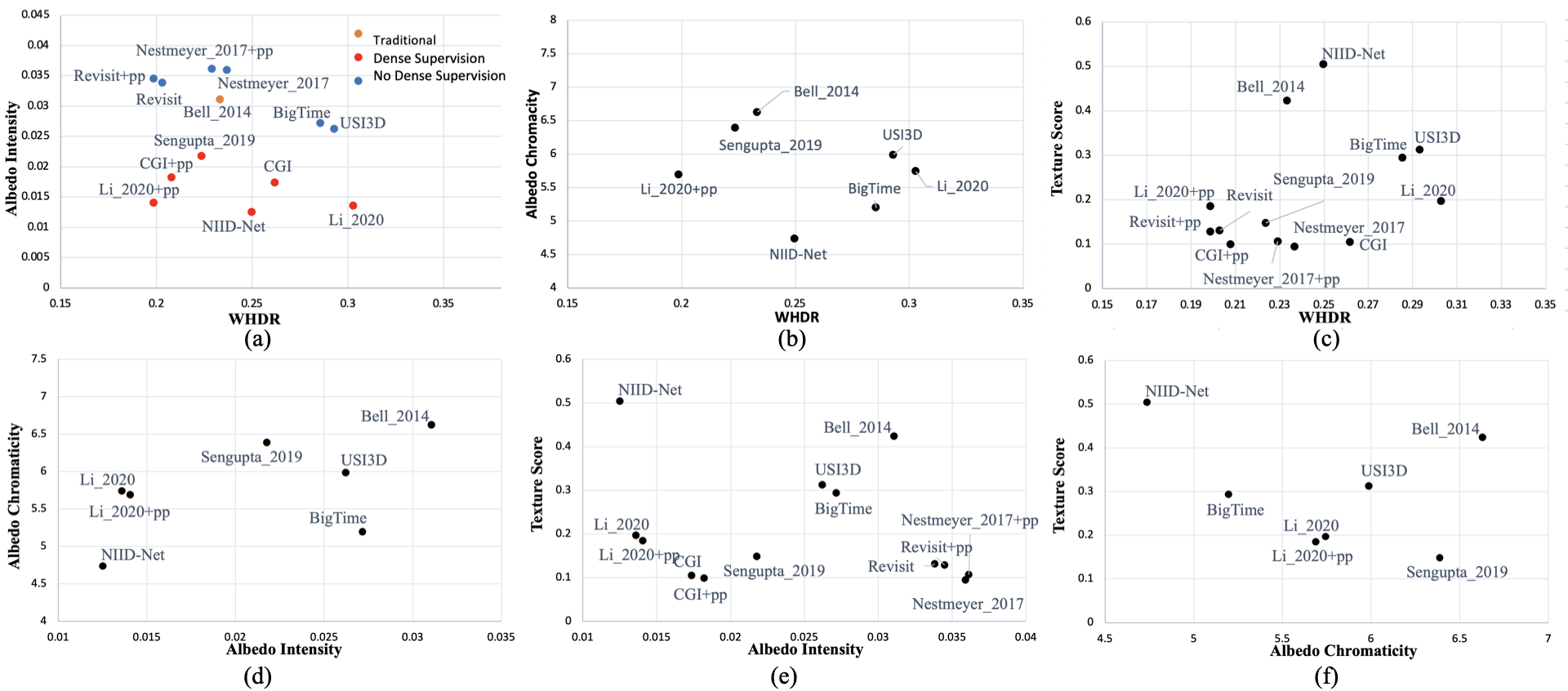}
    \caption{The three charts from left to right are: (a) WHDR vs. albedo intensity error, (b) WHDR vs. albedo chromaticity Error, (c) WHDR vs. Texture Score. In WHDR vs. albedo intensity error, data points with ``+pp'' after the name have post-processing applied. }
    \label{fig:whdr_mean_deltae_tex}
\end{figure*}

\begin{figure*}[!h]
    \centering
    \includegraphics[width=\textwidth]{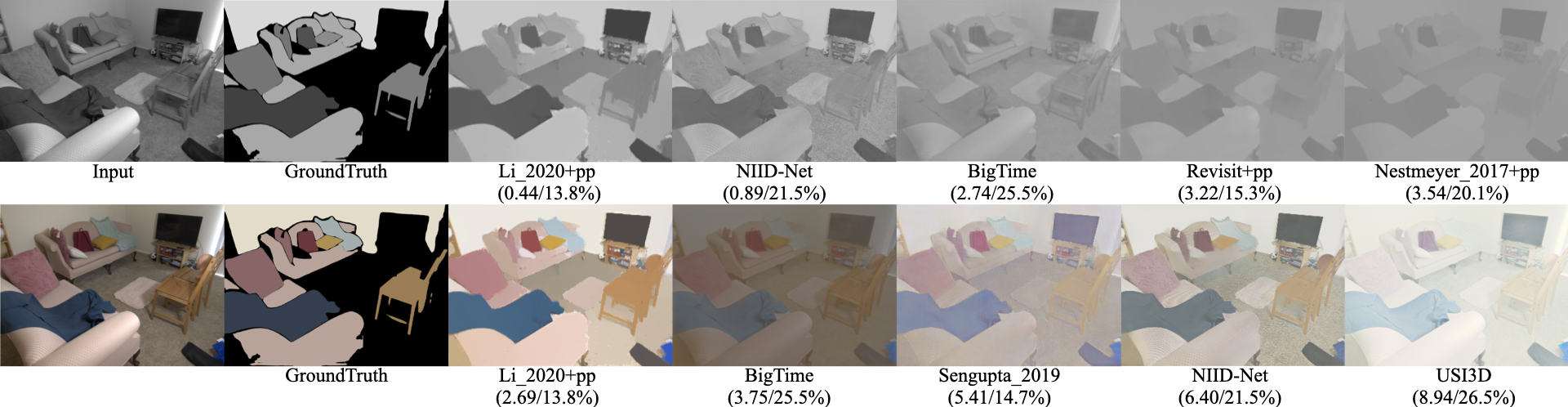}
    \caption{Qualitative comparison of albedo intensity and chromaticity for different algorithms. The number in parenthesis are albedo intensity error ($\times 10^{-2}$)/WHDR (for row 1) and albedo chromaticity error/WHDR (for row 2). ``+pp'' indicates post-processing. Albedos are converted to (grayscale) images in sRGB space for visualization.}
    \label{fig:mse_whdr_qual}
\end{figure*}
\begin{figure*}[!h]
    \includegraphics[width=\textwidth]{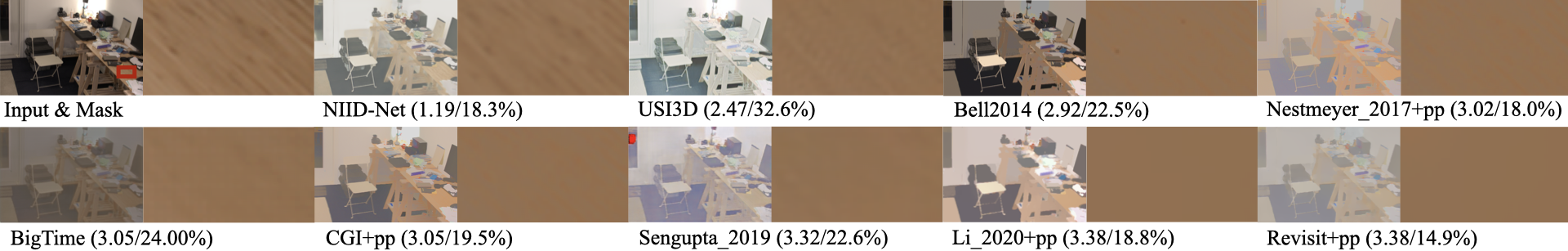}
    \caption{Qualitative comparison of albedo texture. Red bounding box indicates the masked region. Odd columns are albedo images, and even columns are zoomed-in masked region. Numbers in parenthesis are texture ($\times10^{-1}$)/WHDR metrics. Methods with good WHDR do not necessarily perform well on texture metrics.}%
    \label{fig:texture_qual}
\end{figure*}

\noindent\textbf{Albedo Reconstruction Algorithms.} We run 9 algorithms, which are \textbf{Bell\_2014}\cite{iiw}, \textbf{CGI}\cite{cgintrinsics}, \textbf{Nestmeyer\_2017}\cite{Nestmeyer_2017}, \textbf{Revisit}\cite{Fan_2018_CVPR}, \textbf{BigTime}\cite{bigtime}, \textbf{Sengupta\_2019}\cite{nir_2019}, \textbf{Li\_2020}\cite{complex_indoor}, \textbf{USI3D}\cite{usi3d}, and \textbf{NIID-Net}\cite{niidnet}, on our dataset and on the IIW portion of the SAW dataset with officially released code. We evaluated algorithms on albedo intensity, albedo chromaticity, and the WHDR metric and texture metric on our MAW dataset. We report WHDR score on our collected WHDR annotation which is defined on 20\% of each scene.

\newcommand{\indicatorpadding}{\phantom{**}}

\noindent\textbf{Using WHDR conversion function in linear color space:} 
When applying the conversion function $\hat J_{i, \delta}$ introduced in section 2, we use \(R_{1, i}\) and  \(R_{2, i}\) in linear color space, the same as IIW\cite{iiw}. Note that the conversion function \(\hat{J_{i, \delta}}\) considers two points the same if their ratio is smaller than $1+\delta$. This translates to a fixed difference of $\delta$ in log linear albedo space, which roughly corresponds to a fixed difference in human perception\cite{log_perceive}. A fixed ratio does not translate to a fixed perceptual difference in sRGB albedo space. Some authors \cite{paradigms} in previous works reported WHDR with different thresholds. Here we stick to the original evaluation procedure\cite{iiw} and use fixed \(\delta\) of \(0.1\).

Results of evaluating different albedo reconstruction algorithms are shown in Sec. \ref{sec:testing} and Table \ref{tab:metrics}. Then we show how finetuning on  our MAW dataset significantly improves albedo reconstruction of these algorithms (reported for three SOTA algorithms) in Sec. \ref{sec:finetuning} and Table \ref{tab:finetuning}.

\subsection{Evaluating Existing Algorithms}

\label{sec:testing}
We present the results of evaluating different albedo reconstruction algorithms to measure WHDR, intensity, chromaticity and texture metrics on MAW dataset in Table \ref{tab:metrics}. Overall we observe that some algorithms perform better on certain metrics and worse on others. 

In Fig~\ref{fig:iiw_comparison_2}, we show three qualitative examples that highlight this phenomenon. In first row, even though Li\_2020~\cite{complex_indoor} outperforms NIID-Net~\cite{niidnet} in terms of WHDR score and albedo intensity score, Li\_2020 has a warm color tinge and renders the albedo as brown rather than white. On the second row, Revisit~\cite{Fan_2018_CVPR} has a much better WHDR score, but the albedo has very low contrast and a worse albedo intensity score. On the third row, despite that Revisit~\cite{Fan_2018_CVPR} has better WHDR score, all the texture details are smoothed out, and thus has worse texture score than NIID-Net~\cite{niidnet}.

We also present the average relative improvement metric over all algorithms using the formulation proposed in equation \ref{eq:comb}. This measure indicates which algorithms are outperforming other algorithms on average including all the different metrics on albedo. Future researchers developing their own albedo reconstruction algorithm can use this performance measure to calculate relative improvement over existing approaches.

Next, we analyze how different algorithms perform on our proposed metrics (intensity, color, and texture) to show that WHDR on its own is inadequate. Our proposed metrics are complementary to WHDR and necessary for measuring albedo. Results on shading metric are also reported.

In Figure \ref{fig:whdr_mean_deltae_tex}, we plot the scatterplot of WHDR vs Albedo Intensity, WHDR vs Albedo Chromaticity and WHDR vs Texture. WHDR does not seem to strongly correlated with any of the metric we propose.

\textbf{Albedo Intensity.} A qualitative comparison of different methods on the albedo intensity metric is shown in Figure~\ref{fig:mse_whdr_qual} row 1. Some of the previous methods overfit to WHDR score, and overly emphasized the `flatness' and ordinal relationship of the albedo, which damages the absolute intensity of albedo. Nestmeyer\_2017\cite{Nestmeyer_2017} showed lowering input image contrast can significantly improve WHDR score. Revisit, Nestmeyer\_2017, and BigTime all produce very low contrast albedo as shown in Figure~\ref{fig:mse_whdr_qual}, which help them to achieve competitive WHDR score. Such albedos are not physically correct in intensity, and get penalized by our albedo intensity error, as these methods have the worst performance on our albedo intensity error. 

From Figure~\ref{fig:whdr_mean_deltae_tex} and qualitative results shown in Figure~\ref{fig:mse_whdr_qual}, methods (CGI, Sengupta\_2019, Li\_2020, USI3D, NIID-NET) trained on synthetic data with pixel-wise MSE loss (dense supervision) have better albedo intensity score on real data then methods (Revisit, Nestmeyer, BigTime, USI3D) that do not use such supervision (no dense supervision), confirming the importance of synthetic training, but also highlighting the role synthetic data created with physically based rendering play in recovering physically correct albedo. %

\textbf{Albedo Chromaticity.} Figure~\ref{fig:mse_whdr_qual} row 2 shows qualitative comparisons of different methods on albedo chromaticity metric. Nestmeyer\_2017~\cite{Nestmeyer_2017}, CGI~\cite{cgintrinsics}, Revisit~\cite{Fan_2018_CVPR} uses input image chromaticity as the predicted albedo chromaticity. Therefore, we assume they have the performance of using input image chromaticity as albedo chromaticity.  

Even though this white-balanced example contains only a single light source (natural light), many methods still display pronounced overall color artifacts. USI3D~\cite{usi3d} and Sengupta\_2019~\cite{nir_2019} have (blue/green) color tint, while IID-Net\cite{niidnet} have very de-saturated color. 

As shown in table \ref{tab:metrics}, Bell\_2014~\cite{iiw}, Nestmeyer\_2017~\cite{Nestmeyer_2017}, CGI~\cite{cgintrinsics}, and Revisit assumes grayscale lighting. Therefore also perform poorly on albedo chromaticity metric. Again competitive WHDR score does not always imply competitive performance on other metrics. 

Authors of IIW~\cite{iiw} acknowledged that WHDR metric only measures intensity. We found that by evaluating algorithms with WHDR metric alone has led to algorithms which perform poorly on albedo chromaticity, even in simple example such as Figure~\ref{fig:mse_whdr_qual} row 2, highlighting the importance of designing a more comprehensive evaluation framework. As shown in Figure~\ref{fig:iiw_comparison_1}, \ref{fig:iiw_comparison_2}, and \ref{fig:mse_whdr_qual}, we believe there is a significant room for improvement in chromaticity.

\textbf{Albedo Texture.} We use LPIPS to measure texture which is sensitive to image resolution. We try to use resolutions defined in the official code whenever possible.

As shown in Figure \ref{fig:texture_qual}, other than NIID-Net\cite{niidnet}, and maybe USI3D\cite{usi3d}, all methods performed poorly on this example, as barely any wood texture pattern can be seen in the predicted albedos. Such behavior is correctly penalized by our albedo texture metric. As NIID-Net predict perceptually much richer texture detail, our albedo texture metric also assigns significantly higher albedo texture metric. 

Detailed results are in Table \ref{tab:metrics}. Across all examples, we again found all algorithms other than NIID-Net\cite{niidnet} perform poorly on texture metric. Additionally, we found postprocessing (``+pp''), which significantly boost WHDR performance (e.g. for CGI and Li\_2020) seem to further degrade the already poor texture details. Evaluating only on intensity with WHDR metric has led to algorithm design decisions that results in poor albedo texture.
 
\textbf{Shading Metric.} The shading score of all 9 methods are shown in table \ref{tab:metrics}. Since Li\_2020\cite{complex_indoor} and Sengupta\_2019\cite{nir_2019} are inverse rendering algorithms and do not predict shading directly, we use image divide by predicted albedo \(S = I / A\) as shading prediction. With intrinsic decomposition formulation \(I=A*S\), since \(I\) is fixed, given a certain \(A\), \(S\) will also be fully specified. Therefore, performance of various algorithms on shading score closely mirror their performance on albedo (intensity + chromaticity). %

\begin{figure*}
    \includegraphics[width=\textwidth]{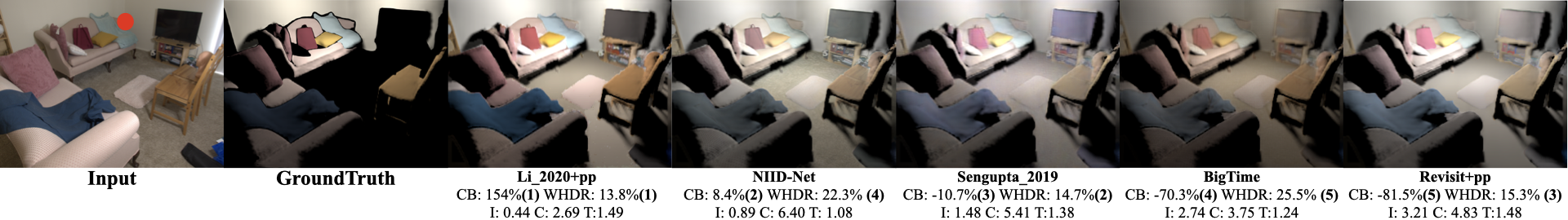}
    \vspace{-2em}
    \caption{
        Perceptual study in downstream tasks and combined metrics. From left to right are input image, relighting result with our measured albedo, relighting results using albedo predicted by 5 algorithms, sorted based on visual quality according to authors (best to worst). CB, I, C, T means combined metrics, albedo intensity, albedo chromaticity and texture respectively. Ranking on combined metrics and WHDR shown in parenthesis after their numbers. We observe that our proposed metrics, individual and combined (CB) are more strongly correlated with the perceptual quality than WHDR.
    }
    \label{fig:relighting} 
\end{figure*} 
\textbf{Validation of metrics in downstream relighting applications.} As shown in Figure \ref{fig:relighting}, to validate our proposed individual and combined metrics, we show qualitative results along with our proposed metrics in downstream relighting task. We use a recent physically based single image relighting technique\cite{indoorlightedit} for relighting. We turn off all light sources in the original image, and insert a lamp whose location is shown by the red sphere. We swap the albedo predicted by the relighting technique\cite{indoorlightedit} with the albedo predicted by the algorithms we are testing. We also remove their indirect illumination module and use only direct illumination, as the indirect illumination module is learned rather than physically based and produces unintuitive and often incorrect results. Our results show that the perceptual quality of downstream tasks like relighting are strongly correlated to our proposed metrics but only weakly correlated with existing WHDR scores.

\newcommand{\alignfttable}{\phantom{(+18.5\%)}}
\begin{table*}[]
    \caption{Finetuning on MAW using WHDR, intensity, and color metrics improves overall performance. Average Relative Improvement is computed over all 6 algorithms using all the metrics, with and without texture. Fixed hyperparameters are used for 4 fold cross validation. Numbers in parenthesis indicates percentage improvement after finetuning compared to before finetuning.}

    \centering
    \begin{tabular}{|c|c|c|c|c|c|c|}
        \hline
    {} & {WHDR (\%)} & {Intensity (\(\times 10^{-2}\))} & {Chromaticity}  & {Texture  (\(\times 10^{-1}\))} & Rel. Improv. (\%) & Shading\\
    {} & Lower better ($\downarrow$) & Lower better ($\downarrow$) & Lower better ($\downarrow$) & Lower better ($\downarrow$) & Higher better ($\uparrow$) & Lower better ($\downarrow$) \\
   \hline
   
   Sengupta\_2019\cite{nir_2019} & 21.7 & 2.03 & 6.37 & 2.40 & -53.0\% & 0.119\\
   + finetune on MAW & 23.8 (-9.7\%) & 1.28 (+58.6\%) & 4.59 (+27.9\%) & 2.34 (+2.5\%) & -7.2\% & 0.106\\
   \hline
   
   Li\_2020\cite{cgintrinsics} & 28.6 & 1.33 & 5.38 & 2.41 & -32.3\% & 0.096\\
   + finetune on MAW & 27.5 (+3.8\%) & 1.00 (+24.8\%) & 4.19 (+22.1\%) & 2.26 (+6.2\%) & +7.3\% & 0.089\\
   \hline
   
   NIID-Net\cite{niidnet} & 24.6 & 1.24 & 4.73 & 1.28 & +29.8\% & 0.106 \\
   + finetune on MAW & 27.6 (+12.2\%) & 0.93 (+25.0\%) & 4.59 (+3.0\%) & 1.07 (+16.4\%) & +55.4\% & 0.092\\
   \hline
\end{tabular}
    \label{tab:finetuning}
\end{table*}
 \begin{figure*}[!h]
    \centering
    \includegraphics[width=0.98\textwidth]{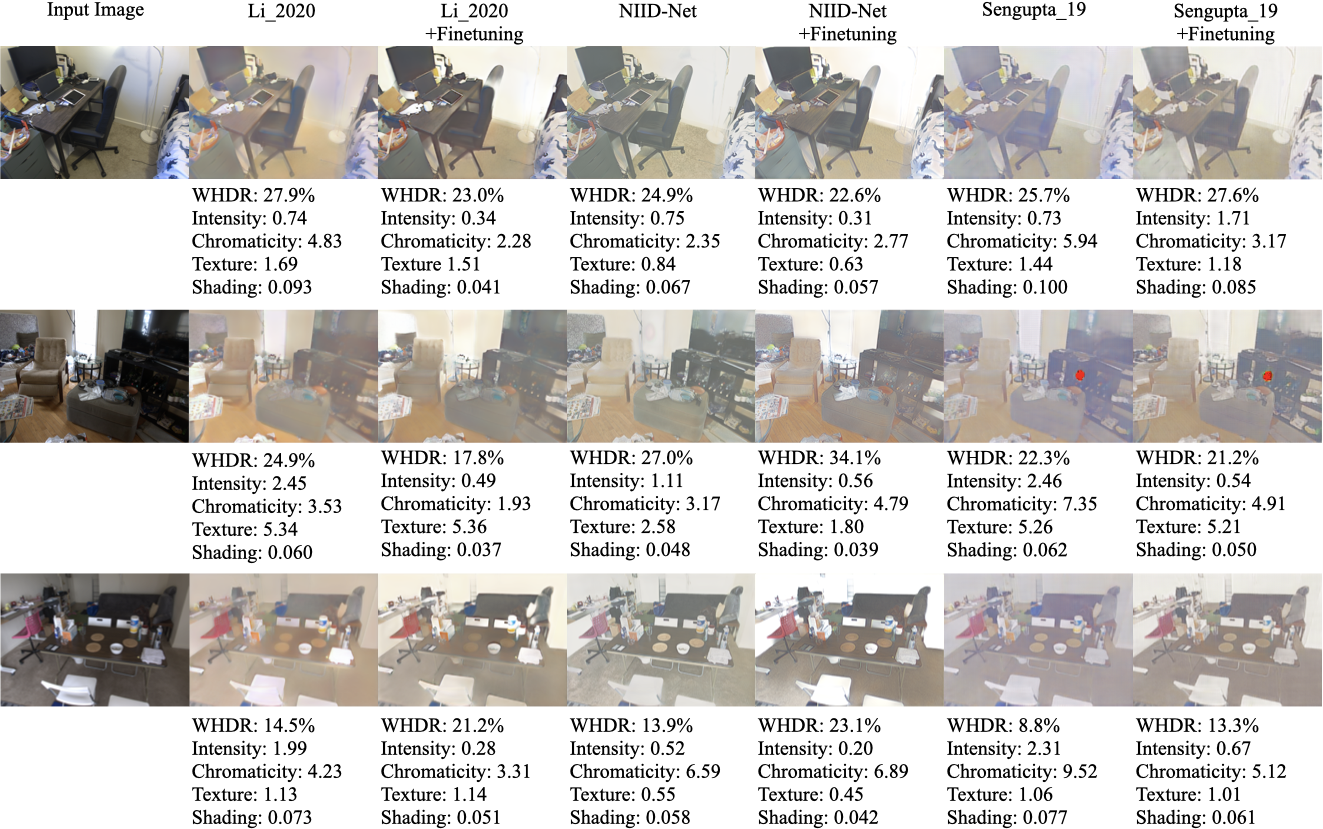}
    \caption{Qualitative comparison of albedo prediction before and after finetuning on the MAW dataset. The scales for Intensity: \(10^{-2}\), Texture: \(10^{-1}\)}
    \label{fig:finetune_qual}
    \vspace{-1.0em}
\end{figure*}
\subsection{Finetuning on MAW}
\label{sec:finetuning}

We pick three SOTA algorithms, Sengupta\_2019\cite{nir_2019}, Li\_2020\cite{complex_indoor}, and NIID-Net\cite{niidnet} for finetuning. We follow the strategy proposed in Sec. \ref{sec:train}.
We perform 4-fold cross-validation on our dataset by splitting our dataset into 4 partitions as equal as possible, subject to the constraint that images from the same building always stay in the same split. A comparison of these algorithms before and after finetuning is shown in Table \ref{tab:finetuning}. 

Sengupta\_2019\cite{nir_2019} and Li\_2020\cite{complex_indoor} do not produce shading image directly. Therefore, for this experiment, we divide the image by each algorithm's respective predicted albedo \(S = I / A\) as the predicted shading image. We also used our own data loading pipeline (e.g., different image resolution), thus the performance differs slightly from the official implementation used in Table \ref{tab:metrics}. However, the setup is consistent throughout the finetuning experiment.

Overall, all three algorithms benefit from the finetuning, achieving higher relative improvement score against others. In Fig \ref{fig:finetune_qual}, we visualize 3 examples before and after finetuning for all three algorithms. Notice how the finetuning has improved most measures. Qualitatively, finetuning has significantly decreased the color tinge of the images. Quantitatively, finetuning improves albedo reconstruction of all three algorithms in intensity and chromaticity and shading and slightly degrade WHDR. 

%% file: content/conclusion.tex
\section{Conclusion}
Estimating and evaluating albedo is challenging due to the lack of ground-truth for in-the-wild images. We show that the current evaluation protocol of measuring WHDR score on IIW dataset is incomplete and thus propose three other measures along with the WHDR score for the complete evaluation of albedo. These measures include albedo intensity, chromaticity, and texture metric on our proposed MAW dataset. We evaluate 9 SOTA algorithms with publicly available codes and we observe that they mostly overfit to the WHDR metric, and neglect other aspects of albedo. 

NIID-Net has the best overall performance in all the algorithms we tested, which surprisingly is not fine-tuned on any real data including IIW~\cite{iiw}. However, NIID-Net do take advantage of a real data pre-trained normal estimation network. Given available real dataset\cite{nyudataset, diode_dataset, replica_dataset}, and encouraging recent results (e.g. normal estimation\cite{aletoric_normal}, depth estimation\cite{iron_depth, dpt}) of indoor geometry estimation, incorporating geometry for physically accurate intrinsic decomposition might be worth exploring for future work.

Overall, we found training with synthetic or real data with physically-based labels to be highly effective in improving albedo reconstruction performance. Particularly with real data, as our finetuning experiments show, even modest amount can significantly improve SOTA algorithms. Future work should further explore using real data with physically-based labels to create better algorithms.

%% file: iccp23_template.bbl
\begin{thebibliography}{10}
\providecommand{\url}[1]{#1}
\csname url@samestyle\endcsname
\providecommand{\newblock}{\relax}
\providecommand{\bibinfo}[2]{#2}
\providecommand{\BIBentrySTDinterwordspacing}{\spaceskip=0pt\relax}
\providecommand{\BIBentryALTinterwordstretchfactor}{4}
\providecommand{\BIBentryALTinterwordspacing}{\spaceskip=\fontdimen2\font plus
\BIBentryALTinterwordstretchfactor\fontdimen3\font minus
  \fontdimen4\font\relax}
\providecommand{\BIBforeignlanguage}[2]{{%
\expandafter\ifx\csname l@#1\endcsname\relax
\typeout{** WARNING: IEEEtran.bst: No hyphenation pattern has been}%
\typeout{** loaded for the language `#1'. Using the pattern for}%
\typeout{** the default language instead.}%
\else
\language=\csname l@#1\endcsname
\fi
#2}}
\providecommand{\BIBdecl}{\relax}
\BIBdecl

\bibitem{1978intrinsics}
H.~Barrow, J.~Tenenbaum, A.~Hanson, and E.~Riseman, ``Recovering intrinsic
  scene characteristics,'' \emph{Comput. vis. syst}, vol.~2, no. 3-26, p.~2,
  1978.

\bibitem{bi_2015}
\BIBentryALTinterwordspacing
S.~Bi, X.~Han, and Y.~Yu, ``An l1 image transform for edge-preserving smoothing
  and scene-level intrinsic decomposition,'' \emph{ACM Trans. Graph.}, vol.~34,
  no.~4, jul 2015. [Online]. Available: \url{https://doi.org/10.1145/2766946}
\BIBentrySTDinterwordspacing

\bibitem{bi_2018}
\BIBentryALTinterwordspacing
S.~Bi, N.~K. Kalantari, and R.~Ramamoorthi, ``Deep hybrid real and synthetic
  training for intrinsic decomposition,'' in \emph{Proceedings of the
  Eurographics Symposium on Rendering: Experimental Ideas \& Implementations},
  ser. SR '18.\hskip 1em plus 0.5em minus 0.4em\relax Goslar, DEU: Eurographics
  Association, 2018, p. 53–63. [Online]. Available:
  \url{https://doi.org/10.2312/sre.20181172}
\BIBentrySTDinterwordspacing

\bibitem{usi3d}
Y.~Liu, Y.~Li, S.~You, and F.~Lu, ``Unsupervised learning for intrinsic image
  decomposition from a single image,'' in \emph{Proceedings of the IEEE/CVF
  Conference on Computer Vision and Pattern Recognition (CVPR)}, June 2020.

\bibitem{Zhou_2015_ICCV}
T.~Zhou, P.~Krahenbuhl, and A.~A. Efros, ``Learning data-driven reflectance
  priors for intrinsic image decomposition,'' in \emph{Proceedings of the IEEE
  International Conference on Computer Vision (ICCV)}, December 2015.

\bibitem{Fan_2018_CVPR}
Q.~Fan, J.~Yang, G.~Hua, B.~Chen, and D.~Wipf, ``Revisiting deep intrinsic
  image decompositions,'' in \emph{Proceedings of the IEEE Conference on
  Computer Vision and Pattern Recognition (CVPR)}, June 2018.

\bibitem{cgintrinsics}
Z.~Li and N.~Snavely, ``Cgintrinsics: Better intrinsic image decomposition
  through physically-based rendering,'' in \emph{Proceedings of the European
  Conference on Computer Vision (ECCV)}, September 2018.

\bibitem{direct_intrinsic}
\BIBentryALTinterwordspacing
T.~Narihira, M.~Maire, and S.~X. Yu, ``Direct intrinsics: Learning
  albedo-shading decomposition by convolutional regression,'' in \emph{2015
  IEEE International Conference on Computer Vision (ICCV)}.\hskip 1em plus
  0.5em minus 0.4em\relax Los Alamitos, CA, USA: IEEE Computer Society, dec
  2015, pp. 2992--2992. [Online]. Available:
  \url{https://doi.ieeecomputersociety.org/10.1109/ICCV.2015.342}
\BIBentrySTDinterwordspacing

\bibitem{shapenet_intrinsics}
J.~Shi, Y.~Dong, H.~Su, and S.~X. Yu, ``Learning non-lambertian object
  intrinsics across shapenet categories,'' in \emph{2017 IEEE Conference on
  Computer Vision and Pattern Recognition (CVPR)}, 2017, pp. 5844--5853.

\bibitem{survey}
\BIBentryALTinterwordspacing
E.~Garces, C.~Rodriguez-Pardo, D.~Casas, and J.~Lopez-Moreno, ``A survey on
  intrinsic images: Delving deep into lambert and beyond,'' \emph{Int. J.
  Comput. Vision}, vol. 130, no.~3, p. 836–868, mar 2022. [Online].
  Available: \url{https://doi.org/10.1007/s11263-021-01563-8}
\BIBentrySTDinterwordspacing

\bibitem{sirfs}
J.~T. Barron and J.~Malik, ``Shape, illumination, and reflectance from
  shading,'' \emph{TPAMI}, 2015.

\bibitem{nir_2019}
S.~Sengupta, J.~Gu, K.~Kim, G.~Liu, D.~W. Jacobs, and J.~Kautz, ``Neural
  inverse rendering of an indoor scene from a single image,'' in
  \emph{Proceedings of the IEEE/CVF International Conference on Computer Vision
  (ICCV)}, October 2019.

\bibitem{complex_indoor}
Z.~Li, M.~Shafiei, R.~Ramamoorthi, K.~Sunkavalli, and M.~Chandraker, ``Inverse
  rendering for complex indoor scenes: Shape, spatially-varying lighting and
  svbrdf from a single image,'' in \emph{Proceedings of the IEEE/CVF Conference
  on Computer Vision and Pattern Recognition (CVPR)}, June 2020.

\bibitem{inverserendernet}
\BIBentryALTinterwordspacing
Y.~Yu and W.~P. Smith, ``Inverserendernet: Learning single image inverse
  rendering,'' in \emph{2019 IEEE/CVF Conference on Computer Vision and Pattern
  Recognition (CVPR)}.\hskip 1em plus 0.5em minus 0.4em\relax Los Alamitos, CA,
  USA: IEEE Computer Society, jun 2019, pp. 3150--3159. [Online]. Available:
  \url{https://doi.ieeecomputersociety.org/10.1109/CVPR.2019.00327}
\BIBentrySTDinterwordspacing

\bibitem{sir_3d}
Z.~Wang, J.~Philion, S.~Fidler, and J.~Kautz, ``Learning indoor inverse
  rendering with 3d spatially-varying lighting,'' in \emph{Proceedings of the
  IEEE/CVF International Conference on Computer Vision (ICCV)}, October 2021,
  pp. 12\,538--12\,547.

\bibitem{openrooms}
Z.~Li, T.-W. Yu, S.~Sang, S.~Wang, M.~Song, Y.~Liu, Y.-Y. Yeh, R.~Zhu,
  N.~Gundavarapu, J.~Shi, S.~Bi, H.-X. Yu, Z.~Xu, K.~Sunkavalli, M.~Hasan,
  R.~Ramamoorthi, and M.~Chandraker, ``Openrooms: An open framework for
  photorealistic indoor scene datasets,'' in \emph{Proceedings of the IEEE/CVF
  Conference on Computer Vision and Pattern Recognition (CVPR)}, June 2021, pp.
  7190--7199.

\bibitem{interiornet18}
W.~Li, S.~Saeedi, J.~McCormac, R.~Clark, D.~Tzoumanikas, Q.~Ye, Y.~Huang,
  R.~Tang, and S.~Leutenegger, ``Interiornet: Mega-scale multi-sensor
  photo-realistic indoor scenes dataset,'' in \emph{British Machine Vision
  Conference (BMVC)}, 2018.

\bibitem{iiw}
S.~Bell, K.~Bala, and N.~Snavely, ``Intrinsic images in the wild,'' \emph{ACM
  Trans. on Graphics (SIGGRAPH)}, vol.~33, no.~4, 2014.

\bibitem{relationship_mid_level}
D.~Zoran, P.~Isola, D.~Krishnan, and W.~T. Freeman, ``Learning ordinal
  relationships for mid-level vision,'' in \emph{2015 IEEE International
  Conference on Computer Vision (ICCV)}, 2015, pp. 388--396.

\bibitem{human_judgement_2015}
T.~Narihira, M.~Maire, and S.~X. Yu, ``Learning lightness from human judgement
  on relative reflectance,'' in \emph{2015 IEEE Conference on Computer Vision
  and Pattern Recognition (CVPR)}, 2015, pp. 2965--2973.

\bibitem{Nestmeyer_2017}
T.~Nestmeyer and P.~V. Gehler, ``Reflectance adaptive filtering improves
  intrinsic image estimation,'' in \emph{Proceedings of the IEEE Conference on
  Computer Vision and Pattern Recognition (CVPR)}, July 2017.

\bibitem{bigtime}
Z.~Li and N.~Snavely, ``Learning intrinsic image decomposition from watching
  the world,'' in \emph{Computer Vision and Pattern Recognition (CVPR)}, 2018.

\bibitem{glosh}
H.~Zhou, X.~Yu, and D.~W. Jacobs, ``Glosh: Global-local spherical harmonics for
  intrinsic image decomposition,'' in \emph{Proceedings of the IEEE/CVF
  International Conference on Computer Vision (ICCV)}, October 2019.

\bibitem{paradigms}
D.~Forsyth and J.~Rock, ``Intrinsic image decomposition using paradigms,''
  \emph{IEEE Transactions on Pattern Analysis \& Machine Intelligence}, no.~01,
  pp. 1--1, oct 2021.

\bibitem{niidnet}
J.~Luo, Z.~Huang, Y.~Li, X.~Zhou, G.~Zhang, and H.~Bao, ``Niid-net: Adapting
  surface normal knowledge for intrinsic image decomposition in indoor
  scenes,'' \emph{IEEE Transactions on Visualization and Computer Graphics},
  2020.

\bibitem{mit_intrinsics}
R.~Grosse, M.~K. Johnson, E.~H. Adelson, and W.~T. Freeman, ``Ground truth
  dataset and baseline evaluations for intrinsic image algorithms,'' in
  \emph{2009 IEEE 12th International Conference on Computer Vision}.\hskip 1em
  plus 0.5em minus 0.4em\relax IEEE, 2009, pp. 2335--2342.

\bibitem{sintel}
D.~J. Butler, J.~Wulff, G.~B. Stanley, and M.~J. Black, ``A naturalistic open
  source movie for optical flow evaluation,'' in \emph{European Conf. on
  Computer Vision (ECCV)}, ser. Part IV, LNCS 7577, {A. Fitzgibbon et al.
  (Eds.)}, Ed.\hskip 1em plus 0.5em minus 0.4em\relax Springer-Verlag, Oct.
  2012, pp. 611--625.

\bibitem{cubam}
\BIBentryALTinterwordspacing
P.~Welinder, S.~Branson, P.~Perona, and S.~Belongie, ``The multidimensional
  wisdom of crowds,'' in \emph{Advances in Neural Information Processing
  Systems}, J.~Lafferty, C.~Williams, J.~Shawe-Taylor, R.~Zemel, and
  A.~Culotta, Eds., vol.~23.\hskip 1em plus 0.5em minus 0.4em\relax Curran
  Associates, Inc., 2010. [Online]. Available:
  \url{https://proceedings.neurips.cc/paper/2010/file/0f9cafd014db7a619ddb4276af0d692c-Paper.pdf}
\BIBentrySTDinterwordspacing

\bibitem{bpb13}
\BIBentryALTinterwordspacing
I.~Boyadzhiev, S.~Paris, and K.~Bala, ``User-assisted image compositing for
  photographic lighting,'' \emph{ACM Trans. Graph.}, vol.~32, no.~4, jul 2013.
  [Online]. Available: \url{https://doi.org/10.1145/2461912.2461973}
\BIBentrySTDinterwordspacing

\bibitem{megadepth}
Z.~Li and N.~Snavely, ``Megadepth: Learning single-view depth prediction from
  internet photos,'' in \emph{Proceedings of the IEEE Conference on Computer
  Vision and Pattern Recognition (CVPR)}, June 2018.

\bibitem{irisformer}
R.~Zhu, Z.~Li, J.~Matai, F.~Porikli, and M.~Chandraker, ``Irisformer: Dense
  vision transformers for single-image inverse rendering in indoor scenes,'' in
  \emph{Proceedings of the IEEE/CVF Conference on Computer Vision and Pattern
  Recognition (CVPR)}, June 2022, pp. 2822--2831.

\bibitem{lighthouse}
P.~P. Srinivasan, B.~Mildenhall, M.~Tancik, J.~T. Barron, R.~Tucker, and
  N.~Snavely, ``Lighthouse: Predicting lighting volumes for spatially-coherent
  illumination,'' in \emph{Proceedings of the IEEE/CVF Conference on Computer
  Vision and Pattern Recognition (CVPR)}, June 2020.

\bibitem{ir_360}
J.~Li, H.~Li, and Y.~Matsushita, ``Lighting, reflectance and geometry
  estimation from 360deg panoramic stereo,'' in \emph{Proceedings of the
  IEEE/CVF Conference on Computer Vision and Pattern Recognition (CVPR)}, June
  2021, pp. 10\,591--10\,600.

\bibitem{hdrplus}
\BIBentryALTinterwordspacing
S.~W. Hasinoff, D.~Sharlet, R.~Geiss, A.~Adams, J.~T. Barron, F.~Kainz,
  J.~Chen, and M.~Levoy, ``Burst photography for high dynamic range and
  low-light imaging on mobile cameras,'' \emph{ACM Trans. Graph.}, vol.~35,
  no.~6, nov 2016. [Online]. Available:
  \url{https://doi.org/10.1145/2980179.2980254}
\BIBentrySTDinterwordspacing

\bibitem{cam_response}
\BIBentryALTinterwordspacing
O.~Burggraaff, N.~Schmidt, J.~Zamorano, K.~Pauly, S.~Pascual, C.~Tapia,
  E.~Spyrakos, and F.~Snik, ``Standardized spectral and radiometric calibration
  of consumer cameras,'' \emph{Opt. Express}, vol.~27, no.~14, pp.
  19\,075--19\,101, Jul 2019. [Online]. Available:
  \url{http://opg.optica.org/oe/abstract.cfm?URI=oe-27-14-19075}
\BIBentrySTDinterwordspacing

\bibitem{rawpy}
{Maik Riechert}, ``rawpy,'' \url{https://github.com/letmaik/rawpy}, 2022.

\bibitem{libraw}
{LibRaw LLC}, ``Libraw,'' \url{https://github.com/LibRaw/LibRaw}.

\bibitem{fbrs}
K.~Sofiiuk, I.~Petrov, O.~Barinova, and A.~Konushin, ``F-brs: Rethinking
  backpropagating refinement for interactive segmentation,'' in
  \emph{Proceedings of the IEEE/CVF Conference on Computer Vision and Pattern
  Recognition (CVPR)}, June 2020.

\bibitem{deltae2000}
\BIBentryALTinterwordspacing
G.~Sharma, W.~Wu, and E.~N. Dalal, ``The ciede2000 color-difference formula:
  Implementation notes, supplementary test data, and mathematical
  observations,'' \emph{Color Research \& Application}, vol.~30, no.~1, pp.
  21--30, 2005. [Online]. Available:
  \url{https://onlinelibrary.wiley.com/doi/abs/10.1002/col.20070}
\BIBentrySTDinterwordspacing

\bibitem{lpips}
R.~Zhang, P.~Isola, A.~A. Efros, E.~Shechtman, and O.~Wang, ``The unreasonable
  effectiveness of deep features as a perceptual metric,'' in \emph{CVPR},
  2018.

\bibitem{Baslamisli_2018}
A.~S. Baslamisli, H.-A. Le, and T.~Gevers, ``Cnn based learning using
  reflection and retinex models for intrinsic image decomposition,'' in
  \emph{Proceedings of the IEEE Conference on Computer Vision and Pattern
  Recognition (CVPR)}, June 2018.

\bibitem{log_perceive}
\BIBentryALTinterwordspacing
L.~R. Varshney and J.~Z. Sun, ``Why do we perceive logarithmically?''
  \emph{Significance}, vol.~10, no.~1, pp. 28--31, 2013. [Online]. Available:
  \url{https://rss.onlinelibrary.wiley.com/doi/abs/10.1111/j.1740-9713.2013.00636.x}
\BIBentrySTDinterwordspacing

\bibitem{indoorlightedit}
Z.~Li, J.~Shi, S.~Bi, R.~Zhu, K.~Sunkavalli, M.~Ha\v{s}an, Z.~Xu,
  R.~Ramamoorthi, and M.~Chandraker, ``Physically-based editing of indoor scene
  lighting from a single image,'' in \emph{ECCV 2022}.

\bibitem{nyudataset}
P.~K. Nathan~Silberman, Derek~Hoiem and R.~Fergus, ``Indoor segmentation and
  support inference from rgbd images,'' in \emph{ECCV}, 2012.

\bibitem{diode_dataset}
\BIBentryALTinterwordspacing
I.~Vasiljevic, N.~Kolkin, S.~Zhang, R.~Luo, H.~Wang, F.~Z. Dai, A.~F. Daniele,
  M.~Mostajabi, S.~Basart, M.~R. Walter, and G.~Shakhnarovich, ``{DIODE}: {A}
  {D}ense {I}ndoor and {O}utdoor {DE}pth {D}ataset,'' \emph{CoRR}, vol.
  abs/1908.00463, 2019. [Online]. Available:
  \url{http://arxiv.org/abs/1908.00463}
\BIBentrySTDinterwordspacing

\bibitem{replica_dataset}
J.~Straub, T.~Whelan, L.~Ma, Y.~Chen, E.~Wijmans, S.~Green, J.~J. Engel,
  R.~Mur-Artal, C.~Ren, S.~Verma, A.~Clarkson, M.~Yan, B.~Budge, Y.~Yan,
  X.~Pan, J.~Yon, Y.~Zou, K.~Leon, N.~Carter, J.~Briales, T.~Gillingham,
  E.~Mueggler, L.~Pesqueira, M.~Savva, D.~Batra, H.~M. Strasdat, R.~D. Nardi,
  M.~Goesele, S.~Lovegrove, and R.~Newcombe, ``The {R}eplica dataset: A digital
  replica of indoor spaces,'' \emph{arXiv preprint arXiv:1906.05797}, 2019.

\bibitem{aletoric_normal}
G.~Bae, I.~Budvytis, and R.~Cipolla, ``Estimating and exploiting the aleatoric
  uncertainty in surface normal estimation,'' in \emph{Proceedings of the
  IEEE/CVF International Conference on Computer Vision (ICCV)}, October 2021,
  pp. 13\,137--13\,146.

\bibitem{iron_depth}
G.~Bae, I.~Budvytis, and C.~Roberto, ``Irondepth: Iterative refinement of
  single-view depth using surface normal and its uncertainty,'' in
  \emph{British Machine Vision Conference (BMVC)}, 2022.

\bibitem{dpt}
R.~Ranftl, K.~Lasinger, D.~Hafner, K.~Schindler, and V.~Koltun, ``Towards
  robust monocular depth estimation: Mixing datasets for zero-shot
  cross-dataset transfer,'' \emph{IEEE Transactions on Pattern Analysis and
  Machine Intelligence (TPAMI)}, 2020.

\end{thebibliography}
